\pdfoutput=1

\documentclass[11pt]{article}

\usepackage[preprint]{acl}

\usepackage{times}
\usepackage{latexsym}
\usepackage{xcolor,colortbl} 
\usepackage[T1]{fontenc}

\usepackage[utf8]{inputenc}

\usepackage{microtype}

\usepackage{inconsolata}

\usepackage{amsmath}
\usepackage{amsfonts}
\usepackage{graphicx}
\usepackage{amssymb}
\usepackage{booktabs} 
\usepackage{musicography}
\usepackage{multirow}
\usepackage[inline]{enumitem} 
\usepackage{graphicx}
\usepackage{subfig}
\usepackage{xcolor}
\definecolor{LightSteelBlue1}{RGB}{202,225,255}
\definecolor{LightPink}{RGB}{245,191,210}
\definecolor{Moccasin}{RGB}{255, 228, 181}
\usepackage{tikz}
\usepackage{pifont}
\usepackage{xspace}
\usepackage{bm}
\usepackage{makecell}
\usepackage{tablefootnote}
\usepackage[nice]{nicefrac}

\usepackage{multirow}

\definecolor{LightSteelBlue1}{RGB}{202,225,255}

\def\method{\textsc{\texttt{G-Tex}}\xspace}
\def\gcn{\textsc{\texttt{Tex-GCN}}\xspace}
\def\gat{\textsc{\texttt{Tex-GAT}}\xspace}
\def\sage{\textsc{\texttt{Tex-SAGE}}\xspace}

\newcommand\blfootnote[1]{%
  \begingroup
  \renewcommand\thefootnote{}\footnote{#1}%
  \addtocounter{footnote}{-1}%
  \endgroup
}

\title{Graph-Guided Post-Hoc Explanations for Self-Rationalization}

\title{Graph-Guided Textual Explanations: Enhancing Faithfulness of Natural Language Explanations with Highlight Cues}

\title{Graph Guidance for Increased Faithfulness of Natural Language Explanations}

\title{Graph-Guided Textual Explanation Generation Framework}

\author{
Shuzhou Yuan$^\ast$$^\clubsuit$,~
Jingyi Sun$^\ast$$^\heartsuit$,~
Ran Zhang$^\diamondsuit$,~
Michael Färber$^\clubsuit$,\vspace{3.5pt}\\
\textbf{
Steffen Eger$^\spadesuit$, ~Pepa Atanasova$^\heartsuit$
~and
Isabelle Augenstein$^\heartsuit$}\vspace{3.5pt}
\smallskip
\\
$^\clubsuit$ScaDS.AI, TU Dresden,~
$^\heartsuit$University of Copenhagen, \\
$^\diamondsuit$University of Mannheim, ~$^\spadesuit$University of Technology Nuremberg\smallskip\vspace{3.5pt}
\\
\texttt{\{jisu, pepa, augenstein\}@di.ku.dk}\\
\texttt{\{shuzhou.yuan, michael.faerber\}@tu-dresden.de}\\
\texttt{ ran.zhang@uni-mannheim.de ~ steffen.eger@utn.de}
}

\begin{document}
\maketitle
\begin{abstract}
Natural language explanations (NLEs) are commonly used to provide plausible free-text explanations of a model's reasoning about its predictions.\blfootnote{$^\ast$ Equal contribution.}
However, recent work has questioned their faithfulness, as they may not accurately reflect the model’s internal reasoning process regarding its predicted answer. In contrast, highlight explanations--input fragments critical for the model's predicted answers--exhibit measurable faithfulness.
Building on this foundation, we propose \textbf{\method}, a \textbf{G}raph-Guided \textbf{T}extual \textbf{Ex}planation Generation framework designed to enhance the faithfulness of NLEs.
Specifically, highlight explanations are first extracted as faithful cues reflecting the model's reasoning logic toward answer prediction. They are subsequently encoded through a graph neural network layer to guide the NLE generation, which aligns the generated explanations with the model’s underlying reasoning toward the predicted answer. Experiments on T5 and BART using three reasoning datasets show that \method improves NLE faithfulness by up to 12.18\% compared to baseline methods. Additionally, \method generates NLEs with greater semantic and lexical similarity to human-written ones.
Human evaluations show that \method can decrease redundant content and enhance the overall quality of NLEs. Our work presents a novel method for explicitly guiding NLE generation to enhance faithfulness, serving as a foundation for addressing broader criteria in NLE and generated text.\looseness=-1


\end{abstract}

\section{Introduction}
\begin{figure}[ht]
    \centering
    \includegraphics[width=1\linewidth]{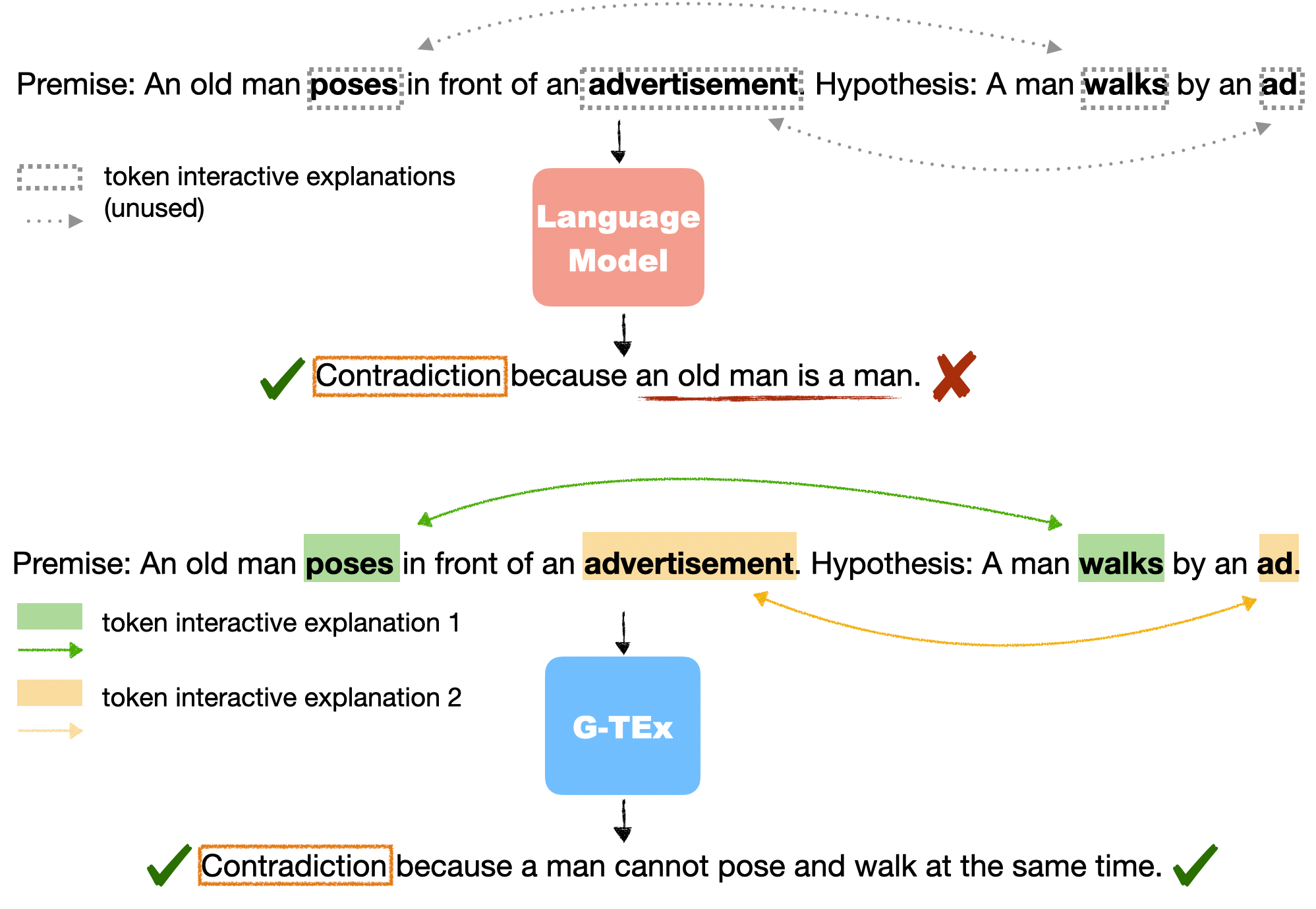}
    \caption{Faithfulness comparison between a self-rationalization model without (top) and with (bottom) the proposed \method. Highlight explanations reveal the model's reasoning behind the predicted label with high faithfulness. Without \method, these important tokens are omitted in the NLE while \method guides the model to incorporate them in the generated NLE.}
    \label{figure:intro_unfaith}
\end{figure}

\begin{figure*}[htbp]
    \centering
    \includegraphics[width=1\textwidth]{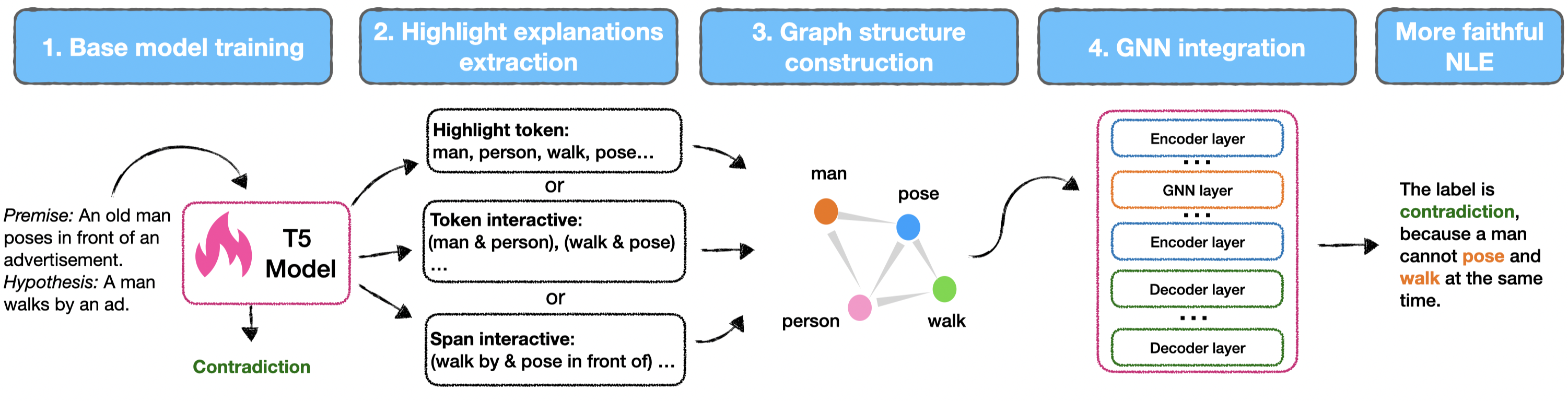}
    \caption{Illustration of our framework \method, which consists of four key steps: (1) We train a base model such as T5 using the task-specific dataset for label prediction (\S\ref{sec:post_hoc_explanation_and_predicted_label:method}). (2) We extract three types of highlight explanations from the trained model (\S\ref{sec:post_hoc_explanation_and_predicted_label:method}). (3) We construct the graph structure based on the highlight explanations (\S\ref{sec:post_hoc_explanation_as_graph:method}) (4) We integrate the graph structure into the model with a GNN layer (\S\ref{sec:Graph_Neural_Network_Layer:method}, \S\ref{sec:integrate_gnn:method}) and fine-tune the overall model for label prediction and NLE generation (\S\ref{sec:self-Rationalization_label:method}).}
    \label{method_overview}
\end{figure*}
Natural Language Explanations (NLEs) produce human-understandable texts to explain the model's prediction process \citep{wiegreffe2021measuring}. Self-rationalization, where the prediction and the corresponding NLE are generated simultaneously, is a commonly used method for NLE generation, which leads to improved agreement between the generated NLE and the produced prediction~\citep{NEURIPS2018_3e9f0fc9,marasovic-etal-2022-shot}. However, existing work~\citep{kumar2020nile,wiegreffe2021measuring} has found that these \textit{NLEs are often unfaithful}, as they may present misleading reasons unrelated to the model’s true decision-making process as illustrated in Figure \ref{figure:intro_unfaith} (top).
This lack of faithfulness undermines the reliability of NLEs in applications where transparency and trust are paramount~\citep{atanasova2023faithfulness,lyu2024towards,parcalabescu-frank-2024-measuring}.


Unlike NLEs, highlight explanations reflect the model's reasoning process by identifying tokens or phrases of the input that are crucial to the model's prediction. 
They can be of three types: \textit{highlight token explanations}, \textit{token interactive explanations} and \textit{span interactive explanations} ~\citep{sun2024unified} (see \S\ref{sec:post_hoc_explanation_and_predicted_label:method} for details).
Though not as plausible as NLEs~\citep{jie2024interpretable}, \textit{the faithfulness of highlight explanations is easy to measure and has been substantially improved in existing works}~\citep{sun2024unified,atanasova-etal-2020-diagnostic}. In this work, we hypothesize that \textit{highlight explanations can be used to improve the faithfulness of NLEs} by using them as explicit cues regarding the important parts of the input that should be present in the generated NLEs. We further hypothesize that as highlight explanations contain concise information about the most important parts of the input, they can further decrease the redundancy of NLEs and improve the overall NLE quality.


Recent efforts to improve the faithfulness of NLEs either rely on external knowledge, crafting prompts or designing the training loss for improving the faithfulness of NLEs directly ~\citep{majumder2021knowledge,marasovic-etal-2022-shot,chuang2024faithlmfaithfulexplanationslarge}. These methods, however, are not targeted at aligning NLEs with a model's inner reasoning but improve their faithfulness only from a model's extrinsic perspective. To address this, and inspired by \citet{yuan-etal-2024-gnnavi} who leverage a Graph Neural Network (GNN) layer to guide the information flow from the input to the generation process, we propose a novel \textbf{G}raph-Guided \textbf{T}extual \textbf{Ex}planation Generation framework (\method) to \textit{enhance the faithfulness of NLEs that allows for explicitly guiding the model’s reasoning with cues derived from the highly faithful highlight explanations}. The graph structure is encoded by a GNN layer, which seamlessly incorporates the highlight explanations into the NLE generation process.
This also allows the model to leverage implicit anchors from the input, improving the generation of explanations.\looseness=-1


As shown in Figure \ref{method_overview}, we first apply a post-hoc attribution method to extract highlight explanations on a fine-tuned model based on its label prediction (\S\ref{sec:post_hoc_explanation_and_predicted_label:method}). Then, we construct a graph with the most important highlight explanations for each instance (\S\ref{sec:post_hoc_explanation_as_graph:method}). A GNN layer is then incorporated to encode the graph within the original self-rationalization model (\S\ref{sec:Graph_Neural_Network_Layer:method}), which is fine-tuned to generate both the final answer prediction and the corresponding NLE simultaneously (\S\ref{sec:self-Rationalization_label:method},\S\ref{sec:integrate_gnn:method}). 

Our findings demonstrate that \method 
substantially improves the faithfulness of NLEs by up to 12.18\% compared to baselines, as evaluated on T5 \citep{raffel2020exploring} and BART~\citep{lewis-etal-2020-bart}(see \S\ref{sec:experimental_setting}) using e-SNLI~\citep{NEURIPS2018_4c7a167b}, ComVE~\citep{wang-etal-2020-semeval} and ECQA ~\citep{aggarwal-etal-2021-explanations} datasets (see \S\ref{sec:results:Faithfulness Evaluation}). Additionally, \method generates NLEs with enhanced semantic and lexical similarity, as evaluated with SacreBLEU~\citep{post-2018-call} and BERTScore~\citep{bert-score} respectively (see \S\ref{sec:automatic_metric_results}). 
Human evaluations further reveal improvements in decreasing redundancy and enhancing the overall quality of the generated NLEs (see details in \S\ref{Human Evaluation}). 
Across the different types of highlight explanations, \textit{token and span interactive explanations} are more effective when the input text involves interaction between different parts. However, when the input consistently includes the same instruction, \textit{highlight token explanations} prove to be more beneficial. Overall, our work introduces a novel method for explicitly guiding the NLE generation to improve faithfulness, serving as a stepping stone for addressing additional criteria for NLE and generated text.

\section{Related Work}
\paragraph{Faithfulness of Natural Language Explanations} 
NLEs are coherent free-text explanations about the reasons behind a model's prediction. Most commonly, NLEs are produced with a self-rationalization set-up where the model generates both a target task prediction and its NLE~\citep{narang2020wt5,tang2021cognitive,atanasova-etal-2020-generating-fact,liu2024is,NEURIPS2023_87e82678,liu-etal-2023-mgr,liu2023decoupled,liu2024enhancing}. As automatically generated NLEs suffer from faithfulness issues~\citep{kumar2020nile,wiegreffe2021measuring,atanasova2023faithfulness,lyu2024towards}, existing work has explored different ways to improve that.
\citet{majumder2021knowledge} propose to first select the important parts of the input, then leverage an external commonsense knowledge generative model
to get commonsense knowledge snippets about these highlights, and finally, use the soft representations of the latter for the NLE generation. 
Another line of work focuses on constructing suitable prompts for NLE generation~\citep{marasovic-etal-2022-shot}. Furthermore, \citet{wang2022pinto} propose to prompt the model to generate the NLE and then fine-tune the LM with a counterfactual regularization loss to make the final prediction based on the generated NLE. \citet{chuang2024faithlmfaithfulexplanationslarge} employ an estimator to provide faithfulness scores for generated NLEs. These scores and the NLEs are appended to the input and iteratively refined until the faithfulness scores converge. However, neither of these works uses direct cues from the more faithful highlight explanation for the model's prediction to guide the NLE generation, which is the novel contribution of this paper. 
Overall, existing work improves NLE faithfulness by resorting to external knowledge, crafting prompts or altering the generation loss. We claim that these constitute extrinsic signals, which do not directly address the NLEs' desiderata to faithfully reflect a model's inner reasoning. Our proposed method \method directly targets this objective by guiding the generation with cues about the most important parts of the input.\looseness=-1

Existing work has also proposed Chain-of-Thought (CoT) explanations, which reveal the model's intermediate reasoning steps before giving its final answer ~\citep{zhang2022automatic}. 
These explanations can be unfaithful as well ~\citep{turpin2024language,jie2024interpretable,lanham2023measuring}. To address this, researchers have leveraged CoT distillation techniques 
to train a more faithful small LM using CoT from the teacher LLM
~\citep{wang2023scott,zhang2024efficient,paul2024making}, or have guided the original LLM to generate multiple reasoning chains and choose the most faithful one
~\citep{li2024towards,jie2024interpretable}.
Notably, we do not focus on the CoT method for generating NLEs, as it requires specialized training data, such as reasoning chains or step-by-step intermediate explanations leading to the final answer. Moreover, CoT views faithfulness as alignment between the generated explanation and the predicted label, which differs from our focus on faithfulness to the model’s internal reasoning process. 

\paragraph{Highlight Explanations for Model Steering} 
Prior works have found that the model's reasoning capability can be enhanced by human-annotated highlight explanations alongside the original input~\citep{wei2022chain,lampinen2022can}.
\citet{krishna2023post} automate the process of filling the extracted highlights into few-shot templates, which enhances model accuracy across tasks such as CommonsenseQA~\citep{talmor-etal-2019-commonsenseqa}. \citet{zhang2024model} propose iterative prompting, where the model first generates a sentence summarizing the input. This sentence is then matched with the most similar sentence from the input, with similarity calculated by an encoder, to refine the prompt and steer the model to produce an answer more accurately. \citet{bhan-etal-2024-self} convert highlight explanations into NLEs using a predefined template, which is then employed to prompt the model for more accurate answers. Though they regard the NLE generation as the intermediate step, the faithfulness of these NLEs is not even evaluated.
In contrast, our approach focuses on enhancing the faithfulness of the generated NLEs by integrating highlight explanations directly into the model architecture to guide NLE generation. 


\paragraph{Graph Neural Networks for Natural Language Processing} Graph neural networks (GNNs) are primarily used for graph-related tasks such as drug discovery \citep{han2021reliable,hu2021opengraphbenchmarkdatasets}. An increasing number of researchers are exploring their potential applications in NLP tasks \citep{yasunaga-etal-2021-qa,fei-etal-2021-iterative,lin-etal-2021-bertgcn}. GNNs have been utilized in tasks like graph-to-text generation \citep{gardent-etal-2017-webnlg, yuan-faerber-2023-evaluating} and graph-enhanced question answering \citep{zhang2022greaselm}, typically encoding complex graph and node representations \citep{koncel-kedziorski-etal-2019-text}. \citet{yuan-farber-2024-grasame} leverage GNNs to encode token-level structural information by modifying the self-attention mechanism in language models. Additionally, \citet{yuan-etal-2024-gnnavi} propose a GNN-based method for information aggregation paired with a parameter-efficient fine-tuning approach. Inspired by previous work,
we 
use GNNs to encode the highlight explanations with high faithfulness to the generation process of NLEs.\looseness=-1



\section{Methodology}

In this section, we provide a detailed overview of \method, as illustrated in Figure \ref{method_overview}. We begin by introducing the self-rationalization model in \S\ref{sec:self-Rationalization_label:method}.
In \S\ref{sec:post_hoc_explanation_and_predicted_label:method}, we describe the training of the base model for label prediction and extracting post-hoc highlight explanations as Steps 1 and 2. In Step 3 and \S\ref{sec:post_hoc_explanation_as_graph:method}, we outline the construction of graph structures. Finally, in Step 4, we present the GNN layer (\S\ref{sec:Graph_Neural_Network_Layer:method}) and explain its integration with language models (\S\ref{sec:integrate_gnn:method}).\looseness=-1

\subsection{Overview: Self-Rationalization Model}
\label{sec:self-Rationalization_label:method}
Self-rationalization models jointly generate the task labels and NLEs to explain their reasoning for the predicted answer \citep{wiegreffe2021measuring}. We frame this as a text-to-text generation task. Note that we are working with tasks containing two separate parts in the input, e.g., a premise and a hypothesis on the e-SNLI dataset (see more details in \S\ref{sec:dataset}).
Given a sequence of tokens $x = (x_1, \ldots, x_{m+n})$ as input, where the first part of the input contains $m$ tokens and the second part $n$ tokens, the model $M$ generates a label $y_0$ and a sequence of tokens for the NLE $y = y_0\oplus(y_1, \ldots, y_{l})$, where $\oplus$ denotes the concatenation of one label token and $l$ NLE tokens.\footnote{See App. \ref{sec:app:esnli comve example} input and output example for e-SNLI.} 
The text generation task, encompassing both label generation and explanation generation, is implemented by a pre-trained LM with a language modeling head on top. Building on this, we insert a graph structure $\mathcal{G}$ into the standard self-rationalization model (LM) to encode the information from the highlight explanations, particularly for interactions between tokens and spans, resulting in our model $M_{G-TEX}$ (see below). We fine-tune this model by minimizing the cross-entropy loss for the target sequence $y$ following the same process of the standard encoder-decoder transformer model. (see Section \ref{sec:integrate_gnn:method} for details on the encoding process after integrating the GNN layer into the self-rationalization model):
\begin{equation}\footnotesize
    \mathcal{L} = -\sum_{i=1}^{|y|} \log P_\phi(y_i | y_{1:i-1}, x, \mathcal{G}),
\end{equation}
where $P_\phi$ is the LM's generative probability.

\subsection{Post Hoc Highlight Explanation and Predicted Label}
\label{sec:post_hoc_explanation_and_predicted_label:method}
As illustrated in Figure \ref{method_overview}, we begin by training a base model, $M_{base}$, designed solely to predict the label of the input text. From this model, we extract three types of highlight explanations from the input following \citet{sun2024unified,choudhury-etal-2023-explaining}. 
These highlights serve as cues revealing the model's reasoning process behind its label predictions.

Given an input instance $x = (x_1, \ldots, x_{m+n})$, each \textit{highlight token explanation} contains one token $x_i$ and its assigned importance score $a_i$; each \textit{token interactive explanation} $(x_i,x_j)$ consists of two interactive tokens from two separate parts of the input respectively, as well as an importance score $a_{ij}$; each \textit{span interactive explanation} is formed of two spans $(span_i,span_j)$, where $span_i=(x_p,\dots,x_{p+l_1})$ and $span_j=(x_q,\dots,x_{q+l_2})$ are from two separate parts of the input respectively, also with an assigned importance score $a_{span_i,span_j}$, where $p, p+l_1 \in [1,m], q, q+l_2\in [m+1, m+n]$.

\paragraph{Highlight Token Explanation Generation.} Interactions between features in LMs are primarily captured through attention mechanisms~\cite {vaswani2017attention}. Previous work shows that highlight explanations extracted by attention-based methods show higher faithfulness than other explainability techniques~\cite{sun2024unified}. Building on this, we use attention weights as the basis for deriving importance scores for all types of highlight explanations.
To retain the unique contributions of individual attention heads -- each designed to focus on specific aspects of the data~\citep{rogers-etal-2020-primer} -- we follow the approach of \citet{choudhury-etal-2023-explaining} to identify the most important attention head for a specific label prediction. We use the final attention layer of the model's decoder, which generates the final token representations used in generation. (see App. \ref{sec:app:post hoc explanation generation} for details). Subsequently, we calculate the importance score $a_i$ for a target token $x_i$ by averaging the self-attention scores assigned to $x_i$ from all other tokens within the input text, following \citet{jain-wallace-2019-attention,sun2024unified}. The extracted \textit{highlight token explanation} set for instance $x$ is noted as $HT=\{(x_i,a_i)|i \in [1, m+n]\}$.  

\paragraph{Token Interactive Explanation Generation.} Using the most important attention head identified as described above, we calculate the importance score $a_{ij}$ for each \textit{token interactive explanation} by averaging the attention weights between these two tokens $x_i$ and $x_j$ following \citet{clark-etal-2019-bert}. The \textit{token interactive explanation} set for instance $x$ is $TI=\{((x_i,x_j),a_{ij})|i\in [1,m],j\in[m+1,n])\}$.

\paragraph{Span Interactive Explanation Generation.}
Since \textit{token interactive explanations} may not convey meaningful information on their own, \citet{choudhury-etal-2023-explaining} suggest using span interactions, which consist of more coherent phrases and are found to be more plausible~\citep{sun2024unified}. 
Following their approach, we apply the Louvain algorithm~\citep{blondel2008fast} to extract \textit{span interactive explanations} by identifying communities of token interactions.
Tokens are treated as nodes, with the importance scores of token pair interactions used as edge weights. The communities of token interactions are selected to have dense intra-span and sparse inter-span interactions. For each $x$, span pairs $(span_i,span_j)$ are extracted, and the importance score $a_{span_i,span_j}$ for each span pair is computed by averaging the importance scores of the constituent token pairs. The set of generated \textit{span interactive explanations} is denoted as $SI=\{(x_{span_i,span_j},a_{span_i,span_j})|span_i=(x_p,\dots,x_{p+l_1}), span_j=(x_q,\dots,x_{q+l_2})\}$. 
The number of generated span pairs depends on the community detection algorithm and is $< m!*n!$ since only neighboring tokens within the same community can form spans, and spans must come from different parts of the input to form valid pairs. 

\subsection{Post Hoc Highlight Explanations as a Graph}
\label{sec:post_hoc_explanation_as_graph:method}

\begin{figure}[!t]
    \centering
    \includegraphics[width=1\linewidth]{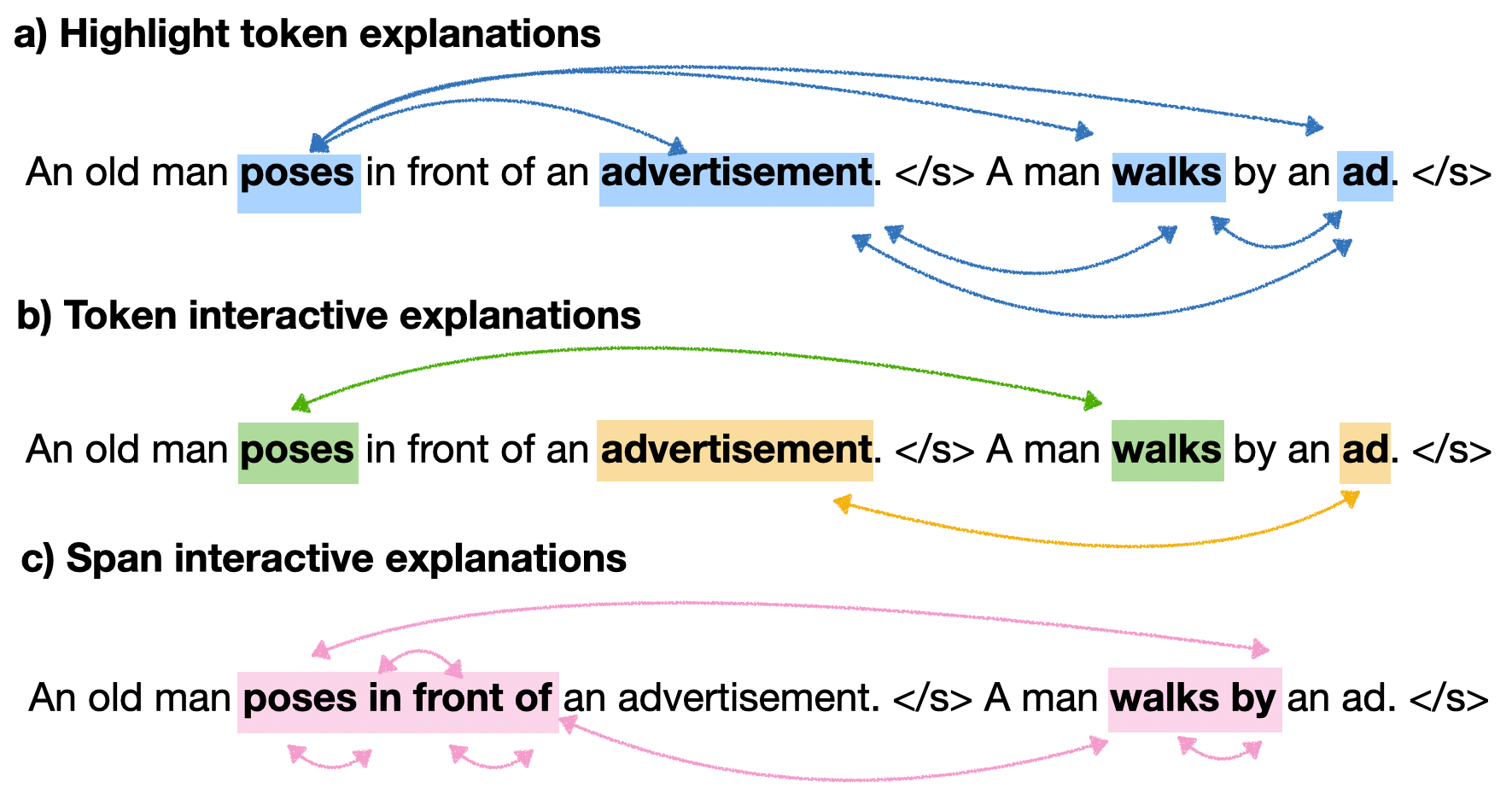}
    \caption{We generate three different types of post-hoc highlight explanations and use them to construct graph structures guiding the NLE generation within our framework. For simplicity, we present only a subset of the explanations for each type.}
    \label{graph_structures}
\end{figure}

We build graph structures based on the three different types of highlight explanations (see Figure \ref{graph_structures}). Notably, we treat each token as a node in the graph structure and assign edges between the extracted tokens. Following \citet{yuan-farber-2024-grasame}, an edge is also assigned to connect the subtokens if a word is tokenized into several subtokens.

\textbf{Highlight Token Explanation} 
We use the importance scores derived in Section \S\ref{sec:post_hoc_explanation_and_predicted_label:method} to select the top-k\% most important highlight token explanations, as less important tokens might introduce noise. Then we assign equally weighted bidirectional edges between these tokens to ensure information flow among them (see Figure \ref{graph_structures}a).


\textbf{Token Interactive Explanations} 
We also select the top-k\% token interactive explanations with the highest importance scores. Then equally weighted bidirectional edges are assigned to connect the tokens within each token interaction (see Figure \ref{graph_structures}b).

\textbf{Span Interactive Explanation}
As only a few spans are extracted from the input text as described in Section \ref{sec:post_hoc_explanation_and_predicted_label:method}, all the interactive spans are used to construct the graph structure. Within a span, all subtokens are connected. Between spans, tokens are connected with each other (see Figure \ref{graph_structures}c).

\subsection{Graph Neural Network Layer}
\label{sec:Graph_Neural_Network_Layer:method}
The GNN layer aggregates information of highlight explanations to model graph and node representations based on the graph structures as introduced in \S\ref{sec:post_hoc_explanation_as_graph:method}.
We define a bidirectional graph $\mathcal{G}$ as a triple $(\mathcal{V}, \mathcal{E}, \mathcal{R})$ with a set of nodes $\mathcal{V} = \{v_1, \ldots, v_n\}$ (one node for each token), a set of relation types $\mathcal{R}$\footnote{We consider only one type of relation: the bidirectional edge between nodes $v$ and $v^{\prime}$, with all edges weighted equally for initialization, note that the edge values will update during fine-tuning}., and a set of edges \( \mathcal{E} \) of the form \( (v, r, v^{\prime}) \) with $v, v' \in \mathcal{V}$, and $r \in \mathcal{R}$.
Each node $v_i$ is associated with a feature vector $h_i$, which represents the hidden states of the $i$-th token in the $l$-th layer. 

The node representations in the GNN layer are updated by aggregating information from neighboring nodes by different aggregation algorithms depending on the chosen GNN architecture. In our work, we employ three most representative and widely used GNN architectures following previous work \citep{yuan-etal-2024-gnnavi,yuan-farber-2024-grasame}: 
Graph Convolutional Network (GCN, \citet{kipf2017semisupervised}), Graph Attention Network (GAT, \citet{velivckovic2018graph}) and GraphSAGE \citep{NIPS2017_5dd9db5e}. While GCN aggregates information from neighboring nodes uniformly, GAT introduces attention weights to prioritize and aggregate incoming information.\footnote{Details of the learning processes for GCN and GAT are provided in App. \ref{appendix:gnn_algo}.} GraphSAGE, on the other hand, incorporates information from the current node and its neighboring nodes as follows:
\begin{equation}\footnotesize
  h_v = \sigma\left(W\left(h_v^{(l)}\oplus\text{AGG}(\{h_{v^{\prime}}^{(l)}, \forall v^{\prime} \in N(v)\}) \right)\right)
\end{equation} 
where $h_v$ denotes the updated node representation of $v$, $h_{v^{\prime}}^{(l)}$ is the token representation of its neighbouring nodes from $l$-th layer, $\sigma$ the activation function, $W$ are the trainable parameters of the GNN, $N(v)$ includes all the neighbouring nodes of $v$. The concatenation function $\oplus$ concatenates aggregated information with the node's current representation, and the aggregation function AGG aggregates the information flowing from the neighboring nodes using techniques such as mean, pool, and LSTM.\footnote{Mean aggregation is applied to GraphSAGE in this work.}

\subsection{Integrating GNN in Language Models}\label{sec:integrate_gnn:method}

As illustrated in Figure \ref{method_overview}, Step 4, we integrate a GNN layer into the LM by stacking it on top of the \( n \)-th encoder layer. \citet{yuan-etal-2024-gnnavi} demonstrated that incorporating a GNN into LLMs is most effective when placed in the last three-quarters of the layers, following the principles of information flow theory \citep{wang-etal-2023-label}. In line with prior work, we similarly position the GNN layer at the \(\nicefrac{3}{4}\)-th encoder layer. The GNN layer takes token representations from the \( l \)-th encoder layer, processes them along with graph structures derived from highlight explanations, and then forwards the augmented representations $h_v$ to the next encoder layer $l+1$, which can be formulated as: 
\begin{equation}\footnotesize
\tilde{h}^{(l)} = \text{LayerNorm}(h_v + \text{Attention}(h_vW^Q, h_vW^K, h_vW^V))
\end{equation}
\begin{equation}\footnotesize
h^{(l+1)} = \text{LayerNorm}(\tilde{h}^{(l)} + \text{FFN}(\tilde{h}^{(l)}))
\end{equation}
The rest of the model architecture remains unchanged.


\section{Experiments}

\subsection{Datasets}
\label{sec:dataset}
We use three widely adopted reasoning datasets with human-annotated explanations: \textbf{e-SNLI}~\citep{NEURIPS2018_4c7a167b}, \textbf{ComVE}~\citep{wang-etal-2020-semeval} and \textbf{ECQA}~\citep{aggarwal-etal-2021-explanations}. \textbf{e-SNLI} extends SNLI with human-annotated explanations for each premise-hypothesis pair, providing both the correct label (entailment, contradiction, or neutral) and a human-annotated NLE for why the label was chosen. \textbf{ComVE} provides natural language explanations identifying which of the two provided statements contradicts common sense. \textbf{ECQA} is a multiple-choice question-answering dataset with human-annotated explanations for each choice.\footnote{In order to explore how different highlight explanations affect faithfulness, we reformulate e-SNLI, ECQA and ComVE into different formats. While the input for e-SNLI and ECQA consists of two distinct sentences, ComVE always includes the same question as the first part of the input (see examples in App. \ref{sec:app:esnli comve example}). This distinction is to explore whether the interaction between the two input parts is significant.}

\begin{table*}[ht]
\centering
\scalebox{0.66}{
\begin{tabular}{@{}c c|cccc|cccc@{}}
\toprule
\multirow{3}{*}{\textbf{Explanation Type}} & \multirow{3}{*}{\textbf{Model}} 
& \multicolumn{4}{c|}{\textbf{e-SNLI}} & \multicolumn{4}{c}{\textbf{ComVE}} \\ 
\cmidrule(lr){3-6}\cmidrule(lr){7-10}
& & \multicolumn{2}{c}{Unfaithfulness(\%$\downarrow$)} & \multicolumn{2}{c|}{Automatic($\uparrow$)}
& \multicolumn{2}{c}{Unfaithfulness(\%$\downarrow$)} & \multicolumn{2}{c}{Automatic($\uparrow$)} \\
& & Counter & Total & SacreBLEU & BERTScore & Counter & Total & SacreBLEU & BERTScore \\
\midrule
\multicolumn{10}{c}{\textbf{T5-based}} \\ 
\midrule
- & \textbf{\texttt{Fine-tuning\textsubscript{base}}} &
  47.70\ \textnormal{\small $\pm$2.31} & 17.68\ \textnormal{\small $\pm$1.94} & 15.430 & 0.894 &
  92.37\ \textnormal{\small $\pm$1.21} & 68.96\ \textnormal{\small $\pm$2.23}
 & 7.634 & 0.876 \\
\midrule
\multirow{2}{*}{Highlight Token} & \textbf{\texttt{Prompt}} &
  43.61\ \textnormal{\small $\pm$2.86} & 14.71\ \textnormal{\small $\pm$1.16} & 15.686 & 0.898 &
  93.25\ \textnormal{\small $\pm$1.19} & 68.90\ \textnormal{\small $\pm$2.61} & 7.592 & 0.876 \\
& \texttt{\textbf{\sage}} (Ours) &
  33.83\ \textnormal{\small $\pm$1.51} & 11.07\ \textnormal{\small $\pm$1.14} & 16.426 & \textbf{0.908} &
  90.53\ \textnormal{\small $\pm$1.40}
 & \textbf{57.48}\ \textnormal{\small $\pm$0.58}
 & \textbf{9.016} & 0.884 \\
\midrule
\multirow{2}{*}{Token Interactions} & \textbf{\texttt{Prompt}} &
  54.36\ \textnormal{\small $\pm$3.11} & 20.60\ \textnormal{\small $\pm$1.81} & 15.478 & 0.898 &
  \textbf{87.39}\ \textnormal{\small $\pm$1.78} & 77.71\ \textnormal{\small $\pm$ 2.06}
 & 7.028 & 0.888 \\
& \texttt{\textbf{\sage}} (Ours) &
  34.27\ \textnormal{\small $\pm$1.63} & 11.00\ \textnormal{\small $\pm$1.66} & \textbf{16.443} & \textbf{0.908} &
  87.47\ \textnormal{\small $\pm$2.21} & 76.94\ \textnormal{\small $\pm$ 2.33}
 & 6.956 & 0.888 \\
\midrule
\multirow{2}{*}{Span Interactions} & \textbf{\texttt{Prompt}} &
 42.86\ \textnormal{\small $\pm$2.20} & 13.19\ \textnormal{\small $\pm$1.95} & 16.031 & 0.899 &
  89.90\ \textnormal{\small $\pm$0.86}  &  79.70\ \textnormal{\small $\pm$2.15}
 & 7.226 & 0.889 \\
& \texttt{\textbf{\sage}} (Ours) &
  \textbf{33.25}\ \textnormal{\small $\pm$2.18} & \textbf{10.08}\ \textnormal{\small $\pm$2.02} & 16.277 & 0.907 &
  89.64\ \textnormal{\small $\pm$0.91} & 76.39\ \textnormal{\small $\pm$3.36}
 & 7.652 & \textbf{0.891} \\
\midrule[1.5pt]
\multicolumn{10}{c}{\textbf{BART-based}} \\ 
\midrule
- & \textbf{\texttt{Fine-tuning\textsubscript{base}}} &
  57.71\ \textnormal{\small $\pm$2.39} & 22.52\ \textnormal{\small $\pm$1.86} & 15.732 & 0.906 &
  91.09\ \textnormal{\small $\pm$1.81} & 70.50\ \textnormal{\small $\pm$1.68}
 & 10.070 & \textbf{0.891} \\
\midrule
\multirow{2}{*}{Highlight Token} & \textbf{\texttt{Prompt}} &
  57.52\ \textnormal{\small $\pm$3.84} & 24.45\ \textnormal{\small $\pm$0.62} & 15.678 & 0.898 &
   90.23\ \textnormal{\small $\pm$2.10} & 68.82\ \textnormal{\small $\pm$2.97}
 & 10.012 & 0.876 \\
& \texttt{\textbf{\sage}} (Ours) &
  \textbf{44.72}\ \textnormal{\small $\pm$4.71} & 14.75\ \textnormal{\small $\pm$2.13} & 16.318 & \textbf{0.909} &
  \textbf{87.91}\ \textnormal{\small $\pm$2.74} & \textbf{58.32}\ \textnormal{\small $\pm$0.81}
 & \textbf{10.552} & 0.884 \\
\midrule
\multirow{2}{*}{Token Interactions} & \textbf{\texttt{Prompt}} &
  47.73\ \textnormal{\small $\pm$3.16} & 19.59\ \textnormal{\small $\pm$1.72} & 15.478 & 0.898 &
  89.80\ \textnormal{\small $\pm$4.54} & 69.43\ \textnormal{\small $\pm$3.14}
 & 7.215 & 0.888 \\
& \texttt{\textbf{\sage}} (Ours) &
 46.88\ \textnormal{\small $\pm$3.34} & 15.68\ \textnormal{\small $\pm$1.75} & \textbf{16.427} & \textbf{0.909} &
  88.15\ \textnormal{\small $\pm$2.47} & 68.08\ \textnormal{\small $\pm$ 2.47}
 & 7.333 & 0.888 \\
\midrule
\multirow{2}{*}{Span Interactions} & \textbf{\texttt{Prompt}} &
  50.98\ \textnormal{\small $\pm$3.72} & 18.34\ \textnormal{\small $\pm$1.70} & 16.027 & \textbf{0.909} &
  95.17\ \textnormal{\small $\pm$1.18}  & 64.35\ \textnormal{\small $\pm$ 0.94}
 & 7.953 & 0.889 \\
& \texttt{\textbf{\sage}} (Ours) &
  45.17\ \textnormal{\small $\pm$3.52} & \textbf{14.64}\ \textnormal{\small $\pm$1.32} & 16.517 & \textbf{0.909} &
  94.29\ \textnormal{\small $\pm$2.57} & 63.76\ \textnormal{\small $\pm$ 2.49} & 7.953 & \textbf{0.891} \\
\bottomrule
\end{tabular}
}
\caption{Overall evaluation results on e-SNLI and ComVE datasets for T5-based and BART-based models, with our \texttt{\textbf{\method}} model using \texttt{\textbf{\sage}}. 
Counter indicates \textit{Counter Unfaith}, Total indicates \textit{Total Unfaith}, with both the mean values and standard deviations reported from 5 runs with different random seeds. The p-values~\cite{wasserstein2016asa} can be found in Appendix \S\ref{sec:appendix:esnli_comve_uncertainty}, Table \ref{tab:uncertainty}. The best performance of each evaluation metric is in bold. See Appendix \S\ref{sec:appendix:ecqa_t5_bart} for results on ECQA dataset and Appendix \S\ref{appendix:overall_gat_gcn_results}, Table \ref{tab:esnli_comve_results_t5_and_bart_gcn_and_gat} for results of our model using \texttt{\textbf{\gat}} and \texttt{\textbf{\gcn}}.
\label{tab:esnli_comve_results_t5_and_bart}}
\end{table*}

\subsection{Experimental Setting}
\label{sec:experimental_setting}
We select two commonly used models for self-rationalization~\citep{raffel2020exploring,narang2020wt5,marasovic-etal-2022-shot,lewis-etal-2020-bart,huang2023chain,yadav2024tox}, \texttt{T5-large} and \texttt{BART-large} as our base models, both of which follow an encoder-decoder architecture. For these models, we insert the graph at the \(\nicefrac{3}{4}\)-th encoder layer. We are not targeting the decoder-only models as they rely solely on the previous token rather than graph embeddings of all tokens for next-token prediction, which limits the guidance of the highlight explanation graphs, and we encourage the modification to apply to decoder-only models for future work (See Limitations).
Our \method is fine-tuned on the training set, with validation performed on the validation set at each epoch. The BLEU score~\citep{papineni-etal-2002-bleu} is used to select the best-performing checkpoint. Further experimental details can be found in App. \ref{sec:appendix:other_experimental_details}.


\subsection{Models}
We use two baselines in our experiments to compare against \method:

\textbf{\texttt{Fine-tuning\textsubscript{base}}} We fine-tune the base models \texttt{T5-large} and \texttt{BART-large} on the training set of e-SNLI and ECQA for self-rationalization.

\textbf{\texttt{Prompt}} 
To incorporate highlight explanations as part of the input, 
we concatenate the template, \textit{``The most important tokens are: token\textsubscript{1}, token\textsubscript{2}, token\textsubscript{3}, ...''} to the end of the input sentence and fine-tune the models accordingly. The important tokens are extracted from the highlight explanations, consistent with the top-k\% tokens used in \method.

\textbf{\texttt{\method}} For our approach, we utilize the encoder-decoder model \texttt{T5-large} and \texttt{BART-large} as the base models and insert a GNN layer after the \(\nicefrac{3}{4}\)-th encoder layer. This GNN layer injects the structured information from the highlight explanations. We experiment with three distinct types of GNN architectures, which we denote as \texttt{\textbf{\gcn}}, \texttt{\textbf{\gat}}, and \texttt{\textbf{\sage}}, representing Graph Convolutional Networks, Graph Attention Networks, and GraphSAGE, respectively (see \S3.4).

\section{Evaluation}

We conduct a comprehensive evaluation of the models, using a faithfulness test, automatic metrics 
and human assessment on multiple dimensions\footnote{The results and analysis of human evaluation are presented in App. \ref{Human Evaluation}}. As for the label predictions, \method achieves results that are better or comparable to the baselines. We report an overview of the label prediction performance in Table \ref{tab:_acc}, App. \ref{appendix:acc}. 

\subsection{Faithfulness Evaluation}
\label{sec:results:Faithfulness Evaluation}
To assess the faithfulness of the generated NLEs, we apply the counterfactual faithfulness test from \citet{atanasova2023faithfulness}. This method involves inserting random adjectives in front of nouns of the original input, resulting in multiple perturbed instances. If the model’s prediction changes, the newly generated NLE should include the inserted word; otherwise, the original NLE is unfaithful as it is potentially misaligned with the model's reasoning. Note that the unchanged label provides no relevant information about the faithfulness of the NLE. See details in App. \ref{sec:appendix:faithfulness_evaluation_details}.


Following \citet{atanasova2023faithfulness}, we apply this test on the e-SNLI, ComVE and ECQA datasets, calculating: 
(1) the percentage of instances where, for at least one altered input, the inserted word does not appear in the new NLE across instances with label change(\textit{Counter Unfaith}); and (2) the proportion of these unfaithful instances across all instances (\textit{Total Unfaith}).

\paragraph{Results} As shown in Tables \ref{tab:esnli_comve_results_t5_and_bart}, our \method\footnote{We select \texttt{\textbf{\sage}} to present the results for \method, as GraphSAGE demonstrates superior performance in modeling text-based graph structures according to previous work \citep{yuan-farber-2024-grasame}. The results of other \method models and the discussion across all GNN variants can be found in App. \ref{appendix:overall_gat_gcn_results}.} 
We present results on e-SNLI and ComVE as representative datasets for NLI and commonsense QA, respectively, and defer ECQA results to App. \ref{sec:appendix:ecqa_t5_bart}. with T5 as the base model leads up to 9.60\% decrease in \textit{Total Unfaithful} on e-SNLI (20.60\% vs. 11.00\% with token interactive explanations) and up to 11.48\% on ComVE (68.96\% vs. 57.48\% with highlight tokens) compared to the \texttt{Fine-tuning\textsubscript{base}} and \texttt{Prompt}. Similarly, \method with BART as the base model leads up to a 9.70\% decrease in \textit{Total Unfaithful} on e-SNLI (24.45\% vs. 14.75\% with highlight explanations) and up to 12.18\% decrease on ComVE (70.50\% vs. 58.32\% with highlight explanations). 
While \method with T5 slightly underperforms the prompt baseline on ComVE with \textit{token interactive explanations}, overall, \textbf{our method outperforms all baselines in counterfactual unfaithfulness and total faithfulness}.

Across the different highlight explanation types, different datasets yield different results. On the e-SNLI dataset, \textit{span interactive explanations} produce more faithful NLEs with T5-based models (10.08\% \textit{Total Unfaith}). 
For the e-SNLI task, the input text consists of two parts, namely the premise and the hypothesis, and interactive explanations between these parts are of paramount importance in indicating the reasoning process of the models. \textbf{Thus, \textit{token interactive }and \textit{span interactive explanations} tend to improve faithfulness more effectively than \textit{highlight token explanations}}. This aligns with previous work showing that these highlight explanations offer higher faithfulness in recovering a model's prediction~\citep{sun2024unified}. 

However, \textit{highlight token explanations} also show significant benefits when the task input consists of the same instruction/first part. As the first part of the input for ComVE is formulated as the same question, the second part of the input becomes especially important in distinguishing the input text for the models. The results on ComVE indicate that \textit{highlight token explanations} yield the lowest \textit{Total Unfaith} for both T5- and BART-based \method (57.48\% and 58.32\%, respectively). \textbf{Thus, \textit{highlight token explanations} can improve the faithfulness when the interaction between parts of the input is less critical.} 

Our findings demonstrate that while all highlight explanations are significantly important, their utility depends on the task. When the input text involves interaction between different parts, \textit{token and span interactive explanations} are more useful. However, when the input consistently includes the same instruction, \textit{highlight token explanations} are more effective. Nonetheless, regardless of the task, the results again verify that \method effectively leverages different types of highlight explanations for NLE generation, leading to more faithful NLEs. 





\subsection{Automatic Metrics for Similarity between NLEs and Golden explanations}
\label{sec:automatic_metric_results}
To assess the alignemnt of generated NLEs with human-written ones, we measure the similarity between them and the golden human-annotated explanations. A similarity with human-written explanations is used in existing work to indicate how plausible the generated NLEs would appear to end users \cite{sun2024unified}. We employ automatic evaluation metrics \textbf{SacreBLEU}~\citep{post-2018-call} and \textbf{BERTScore}~\citep{bert-score} to capture both lexical and semantic similarity.\footnote{In addition to SacreBLEU and BERTScore, results for other automatic metrics are provided in App. \ref{sec:appendix:t5_other_automatic_metrics}.}

As shown in Table \ref{tab:esnli_comve_results_t5_and_bart}, the automatic evaluation results demonstrate that \method generates NLEs of higher alignment with human-written explanations in terms of lexical and semantic similarity on the e-SNLI dataset, outperforming the \texttt{Fine-tuning\textsubscript{base}} and \texttt{Prompt}. Across all explanation types, \method consistently achieves higher SacreBLEU scores, such as 16.443 for \method with the \textit{token interactive explanation} setting, and better BERTScores, such as 0.909 across most BART-based methods. Regarding the ComVE dataset, \method also generates NLEs with higher SacreBLEU and BERTScore. For BART-based \method, the highest ScareBLEU is 10.552 achieved with \method with \textit{highlight token explanations}. 
\textbf{These results demonstrate that our models generate explanations with improved alignment with human explanations}. Furthermore, they confirm that interactive explanations are more effective for e-SNLI, while highlight token explanations are more beneficial for ComVE, due to the distinct structure of their inputs.

\section{Conclusion}

In this work, we propose \method, a novel framework that incorporates the reasoning process of models to enhance faithfulness in NLEs. \method allows for integrating various types of highlight explanations through a GNN layer within language models.
Evaluated via faithfulness tests, automatic metrics, and human evaluation on three reasoning datasets, \method demonstrates consistent improvements in faithfulness, alignment with human-annotated explanations, and reduced redundancy. 
Our results show that the benefits of different highlight explanations depend on task formulation: \textit{token and span interactive explanations} work best for tasks requiring input interaction, while \textit{highlight token explanations} are more effective when interactions are less critical.
These findings highlight the potential of \method as an interpretable framework that embeds the reasoning process of language models as a graph structure to improve model faithfulness. Future work could explore various graph structures and explanation types to further enhance the versatility and effectiveness of \method for larger models.

\section*{Limitations}
Our work proposes a novel graph-guided framework for natural language explanation generation, utilizing highlight explanations in the form of highlight tokens, token interactives, and span interactives. 
While \method improves the models' faithfulness constantly, we acknowledge several limitations in our approach.

Firstly, we applied \method exclusively to encoder-decoder models. This choice was made not only because encoder-decoder models are better suited for text-to-text format tasks, but also because the encoder is able to embed the graph structure and utilize it to generate each individual token. 
While our approach is potentially applicable to decoder-only models, their architectural differences introduce notable complexities. 
In decoder-only models, token generation relies solely on the hidden states of the preceding token. As a result, significant adjustments, such as carefully integrating text embeddings with graph embeddings, would be required to adapt our method for use with decoder-only models to generate tokens, which is thus outside the scope of this work. Due to limited computational resources, we chose T5-large and BART-large as the models to fine-tune for NLE generation. Their established reasoning capabilities and relatively lightweight nature make them well-suited for our experimental setup. We encourage future work to explore how model scalability affects the quality of generated NLEs.


Secondly, while \method leverages the reasoning process of the models and offers a more transparent and interpretable framework, the internal mechanisms of the GNN layer remain unexplored in this study. Moreover, we use specific graph types to construct the highlight explanations, assigning equal weights to the edges between nodes. Future work could explore weighted edges and alternative graph structures to encode highlight explanations.

Thirdly, while we choose the attention-based methods as the foundation to extract highlight explanations due to their higher faithfulness on ECQA and e-SNLI dataset~\citet{sun2024unified}, it is important to acknowledge other important explainability techniques, such as perturbation-based attribution e.g., Shapley~\citep{lundberg2017unified}), Integrated Gradients~\citep{sundararajan2017axiomatic,serrano-smith-2019-attention} and Saliency Map~\citep{feldhus2022constructing}. It is worth exploring how the highlight explanations generated by different explainability techniques impact the quality of generated NLEs on broader datasets. We leave this exploration for future work.

Lastly, we evaluate the quality of NLEs generated by our model using three reasoning datasets, e-SNLI (NLI task), ComVE and ECQA (commonsense QA task). 
As more datasets meeting these criteria become accessible in the future, we encourage further exploration of our method in additional domains.



\section*{Acknowledgments}

$\begin{array}{l}\includegraphics[width=1cm]{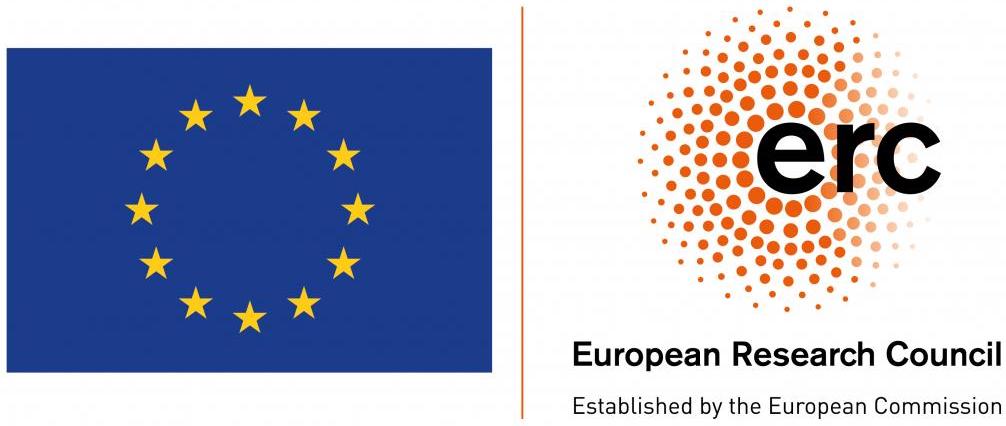} \end{array}$ 
This research was co-funded by the European Union (ERC, ExplainYourself, 101077481), by the Pioneer Centre for AI, DNRF grant number P1, as well as by The Villum Synergy Programme. Views and opinions expressed are however those of the author(s) only and do not necessarily reflect those of the European Union or the European Research Council. Neither the European Union nor the granting authority can be held responsible for them. 

\bibliography{anthology,custom}


\appendix

\section{Post Hoc Explanation Generation Details}
\label{sec:app:post hoc explanation generation}

For each attention head $j$ regarding generating token $k$, When the contribution of input token $i$ $c_{ji}$ is positive, the larger the weight $w_{ji}$, the more important of input token $i$ to $k$. We aggregate the importances for generating $k$ from all input tokens in attention head $j$ as the indication of the overall importance of attention head $j$. 

\section{Raw Running Time of Exacting Highlight Explanations}\label{sec:appendix:time_extract_explanation}

We report the raw running time for extracting highlight explanations on the test set of e-SNLI using T5-based model in Table \ref{tab:time_cost}. Although the span interactive explanation has the longest runtime, it only requires 14 ms to extract explanations for an instance with the longest token range. While extracting explanations adds some computational time, it is not prohibitive for practical use.


\begin{table*}[h]
    \centering
    \scalebox{0.9}{
    \begin{tabular}{lccc}
        \toprule
        \textbf{Explanation Type} & \textbf{[5, 20) Tokens} & \textbf{[20, 40) Tokens} & \textbf{[40, 69] Tokens} \\
        \midrule
        \multicolumn{4}{c}{\textbf{Average Time Cost per Instance (ms)}} \\
        \midrule
        Highlight Token Explanation  & 0.7382  & 1.2903  & 2.5249  \\
        Token Interactive Explanation & 0.3924  & 0.8071  & 1.8725  \\
        Span Interactive Explanation  & 2.5501  & 5.8975  & 14.3293 \\
        \midrule
        \multicolumn{4}{c}{\textbf{Number of Instances in Each Token Range}} \\
        \midrule
        & 4,546  & 5,068  & 192  \\
        \bottomrule
    \end{tabular}}
    \caption{Average time cost (in milliseconds) and instance counts across different token length ranges for three types of highlight explanation extraction using a T5-based model on the e-SNLI test set.}
    \label{tab:time_cost}
\end{table*}

\section{Example of Self-Rationalization Task}\label{sec:app:esnli comve example}

One example of the e-SNLI dataset~\citep{NEURIPS2018_4c7a167b} will be:
\begin{quote}
\raggedright
\textit{
\textbf{Input}: "Premise: A woman is asleep at home. Hypothesis: A woman with a red scarf is giving a shushing sign to the camera in front of shelves of books."\\
\textbf{Output}: "Contradiction. The woman cannot be giving a sign and asleep at the same time."
}
\end{quote}

where \textit{Contradiction} is the predicted label, and the text explains the reason for the prediction.

And we reformulate ComVE dataset~\citep{wang-etal-2020-semeval} as:
\begin{quote}
\raggedright
\textit{
\textbf{Input}: "Which statement of the two is against common sense? \\
1. when it is hot humidity forms \\
2. when it rains humidity forms amaranthine"\\
\textbf{Output}: "2. Water makes humidity, not temperature."
}
\end{quote}

where \textit{2} is the index of the prediction, and the text explains the reason why it is against common sense.

Meanwhile, one example of ECQA~\citep{aggarwal-etal-2021-explanations}:
\begin{quote}
\raggedright
\textit{
\textbf{Input}: "The student was contemplating the problem, that's when he made the what that led him to the answer? \\
action, discovery, reflection, deciding, thinking\\
\textbf{Output}: "discovery. Contemplating on the problem, the student made the discovery or finding that led him to the answer. Contemplating and thinking deeply about the problem may or may not lead to action. Reflection is contemplating of thinking about oneself and not the problem. Deciding is contemplating choice and wrong decisions don’t lead to answer. Thinking won’t necessarily lead to the answer."
}
\end{quote}

where \textit{discovery} is the predicted answer, and the text explains the reason why it is correct and why the others are wrong.

\section{Aggregation Algorithms of GCN and GAT}\label{appendix:gnn_algo}

The learning process of GCN is formulated as:
\begin{equation}\footnotesize
  h_v = \sigma\left(W\sum_{v^{\prime} \in N(v)} \frac{h_{v^{\prime}}^{(l)}}{\lvert N(v)\rvert}\right)
\end{equation} 

where $h_v$ denotes the updated node representation of $v$, $h_{v^{\prime}}^{(l)}$ is the token representation of its neighbouring nodes from $l$-th layer, $\sigma$ the activation function, $W$ are the trainable parameters of the GNN, $N(v)$ includes all the neighbouring nodes of $v$. 

Unlike the average over all neighbouring nodes in GCN, GAT learns an attention weight $\alpha$ for every neighbouring node:
\begin{equation}\footnotesize
  h_v = \sigma\left(\sum_{v^{\prime} \in N(v)} \alpha_{vv^{\prime}} Wh_{v^{\prime}}^{(l)}\right)
\end{equation} 

\section{Performance for Label Prediction}\label{appendix:acc}

We present the performance of all baselines and \method for the label prediction task in Table \ref{tab:_acc}. \method consistently outperforms the baselines on both the e-SNLI and ECQA datasets.

As shown in Table \ref{tab:_acc}, we present our \method models’ performance in answer prediction, where the GNN layer is jointly fine-tuned with the base model alongside all baseline models. It is evident that the \method model achieves better or comparable accuracy to the baseline models, ensuring that \method does not sacrifice answer accuracy while increasing NLE faithfulness.

\begin{table}[ht]
\centering
\scalebox{0.65}{
\begin{tabular}{c c c c c}
\toprule
\multicolumn{2}{c}{\textbf{Method}} & \textbf{Acc\textsubscript{e-SNLI}} & \textbf{Acc\textsubscript{ECQA}} & \textbf{Acc\textsubscript{ComVE}} \\
\midrule
\multicolumn{5}{c}{\textbf{T5-large}} \\
\midrule
\multicolumn{2}{c}{\textbf{\texttt{Fine-tuning\textsubscript{base}}}} & 84.50 & 61.56 & 89.92\\
\cmidrule(lr){1-5}
\multirow{4}{*}{Highlight Tokens} & \textbf{\texttt{Prompt}} & 86.16 & 60.98 & 88.05\\
 & \textbf{\texttt{\gcn}} & 89.79 & 59.87 & 90.86\\
 & \textbf{\texttt{\gat}} & 89.42 & 60.22 & 91.08\\
 & \textbf{\texttt{\sage}} & 89.78 & 60.37 & 92.43\\
\cmidrule(lr){1-5}
\multirow{4}{*}{Token Interactions} & \textbf{\texttt{Prompt}} & 86.02 & 57.17 & 90.48\\     
 & \textbf{\texttt{\gcn}} & 89.88 & \textbf{62.23} & 90.97\\
 & \textbf{\texttt{\gat}} & 89.93 & 61.76 & 90.14\\
 & \textbf{\texttt{\sage}} & 89.94 & 61.25 & 89.76\\ 
\cmidrule(lr){1-5}
\multirow{4}{*}{Span Interactions} & \textbf{\texttt{Prompt}} & 88.92 & 59.14 & 88.14\\
 & \textbf{\texttt{\gcn}} & 89.76 & 59.62 & 89.06\\
 & \textbf{\texttt{\gat}} & 89.10 & 59.02 & 90.36\\
 & \textbf{\texttt{\sage}} & 89.98 & 58.62 & 89.76\\
\midrule
\multicolumn{5}{c}{\textbf{BART-large}} \\
\midrule
\multicolumn{2}{c}{\textbf{\texttt{Fine-tuning\textsubscript{base}}}} & 85.29 & 56.91 & 91.57\\
\cmidrule(lr){1-5}
\multirow{4}{*}{Highlight Tokens} & \textbf{\texttt{Prompt}} & 81.55 & 42.21 & 91.47\\
 & \textbf{\texttt{\gcn}} & \textbf{91.04} & 41.82 & 92.17\\
 & \textbf{\texttt{\gat}} & 90.60 & 50.50 & 92.15\\
 & \textbf{\texttt{\sage}} & 91.03 & 52.73 & \textbf{92.67}\\ 
\cmidrule(lr){1-5}
\multirow{4}{*}{Token Interactions} & \textbf{\texttt{Prompt}} & 90.42 & 54.59 & 90.48\\
 & \textbf{\texttt{\gcn}} & 90.18 & 58.02 & 91.76\\
 & \textbf{\texttt{\gat}} & 89.52 & 55.50 & 86.51\\
 & \textbf{\texttt{\sage}} & 89.44 & 52.46 & 91.77\\ 
\cmidrule(lr){1-5}
\multirow{4}{*}{Span Interactions} & \textbf{\texttt{Prompt}} & 90.35 & 56.38 & 89.13\\
 & \textbf{\texttt{\gcn}} & 90.91 & 51.53 & 91.77\\
 & \textbf{\texttt{\gat}} & 91.03 & 56.94 & 91.06\\
 & \textbf{\texttt{\sage}} & 90.79 & 44.41 & 92.17\\
\bottomrule
\end{tabular}
}
\caption{Overview of model accuracy on e-SNLI, ECQA and ComVE datasets. \method achieves results that are better or comparable to the baselines. The best performance of each evaluation metric across all models is highlighted in bold.}
\label{tab:_acc}
\end{table}

\section{Experimental Details}
\label{sec:appendix:other_experimental_details}
The number of incorporated GNN layers is 1. Final results are reported on the test set with beam search set to 3. We set $k=30$ to take the top 30\% most important highlight explanations. Training is conducted on four NVIDIA A100-SXM4-40GB GPUs, utilizing AdamW \citep{loshchilov2018decoupled} as the optimizer. The learning rate is set to \texttt{3e-4} for both the baselines and \method after grid search. And beam search is set to 3 for the text generation. We use the original train, dev, and test splits for model fine-tuning across all the datasets.

\section{Model Size}\label{sec:appendix:model_size}

Table \ref{tab:num_params} shows the number of trainable parameters comprising the baselines and G-Tex, as well as the training time for one epoch under the same configuration (batch size, optimizer, learning rate, etc.). Notably, the model incorporating GNNs only has approximately up to 0.28\% more parameters than the baseline models T5 and 0.24\% more parameters than the baseline models BART. Overall, the training time for different methods varies by only a few seconds. 
\begin{table}[h]
    \centering
    \footnotesize
    \begin{tabular}{llll}
        \hline
        \textbf{Method} & \textbf{Param\textsubscript{T5}} & \textbf{Param\textsubscript{BART}} & \textbf{Time\textsubscript{T5}} \\
        \hline
        Fine-tuning & 737M  & 406M & 13:51\\
        Prompt      & 737M  & 406M & 14:23\\
        Tex-GCN     & 738M  & 407M & 13:41\\
        Tex-GAT     & 738.1M & 407M & 13:42\\
        Tex-SAGE    & 739.1M & 407M & 13:49\\
        \hline
    \end{tabular}
    \caption{Number of parameters and training time for different methods using T5 and BART.}
    \label{tab:num_params}
\end{table}

\section{Faithfulness Evaluation Method}
\label{sec:appendix:faithfulness_evaluation_details}
Following ~\cite{atanasova2023faithfulness}, we conduct the counterfactual evaluation to assess the faithfulness of the generated NLEs. Specifically, given an input instance $x$ with the model’s original answer $y_0$ and its corresponding NLE tokens $[y_1, \dots, y_{l}]$ (see \S\ref{sec:self-Rationalization_label:method}), we insert a word $x_c$ into $x$, forming a new input $x'$. To ensure the coherence of $x'$,  we only insert random adjectives before nouns. For each original input $x$, we generate candidate insertions at 4 random positions, with 4 candidates per position, resulting in 16 perturbed inputs $x'$ for each instance.  If the model’s prediction changes ($y_0' \neq y_0$), the newly generated NLE should include the inserted word, i.e., $x_c \in [y_1', \dots, y_{p+q}']$; otherwise, the original NLE is unfaithful as it is potentially misaligned with the model's reasoning. Note that the unchanged label provides no relevant information about the faithfulness of the NLE.

\section{Overall Explanation Evaluation Results on e-SNLI and ComVE Dataset of \textbf{\texttt{\method}} using \textbf{\texttt{\gat}} and \textbf{\texttt{\gcn}}}
\label{appendix:overall_gat_gcn_results}

As shown in Table \ref{tab:esnli_comve_results_t5_and_bart_gcn_and_gat}, we also report the results of our models \textbf{\texttt{\method}} using  \textbf{\texttt{\gat}} and \textbf{\texttt{\gcn}}. 

Regarding faithfulness, almost all of our models outperform all the baseline models on both datasets, achieving improvements of up to 17.18\% with the T5-based \textbf{\texttt{\gcn}} on the ComVE dataset,
which demonstrates our approach's effectiveness in enhancing the faithfulness of NLEs. 

Across different highlight explanation types, \textbf{\textit{token interactive explanations}} consistently achieve the best faithfulness results on the e-SNLI dataset, regardless of the base model architecture. In contrast, on the ComVE dataset, \textbf{\textit{highlight token explanations}} consistently demonstrate the highest faithfulness, highlighting the influence of dataset characteristics on the advantages of different explanation types in enhancing NLE faithfulness. For example, on the ComVE dataset, where the first part of the input is a general question in which the statement of the two is against comment sense, the simple interaction between the tokens/spans from the question and the statements might be less informative than simply selecting the important tokens from the statements. \textbf{This suggests that the choice of highlight explanation types to enhance NLE quality, particularly in terms of faithfulness, should be carefully tailored to the specific characteristics of the dataset.
}

Regarding the similarity between the generated NLEs and the golden ones, as measured by automatic metrics, all the NLEs generated by our method on both datasets achieve equal or higher performance than the baselines. Among the different highlight explanation types, NLEs guided by \textbf{\textit{highlight token explanations}} most frequently achieve the highest similarity with the golden ones, both lexically and semantically.

Among the different GNN variants of our \textbf{\texttt{\method}} method, \textbf{\texttt{\gat}}, \textbf{\texttt{\gcn}}, and \textbf{\texttt{\sage}}, there is no consistent trend indicating that any particular GNN layer consistently outperforms the others in improving the faithfulness or the similarity of the NLEs to the golden explanations. 

\begin{table*}[ht]
\centering
\scalebox{0.71}{
\begin{tabular}{@{}c c|cccc|cccc@{}}
\toprule
\multirow{3}{*}{\textbf{Explanation Type}} & \multirow{3}{*}{\textbf{Model}} 
& \multicolumn{4}{c|}{\textbf{e-SNLI}} & \multicolumn{4}{c}{\textbf{ComVE}} \\ 
\cmidrule(lr){3-6}\cmidrule(lr){7-10}
& & \multicolumn{2}{c}{Unfaithfulness(\%$\downarrow$)} & \multicolumn{2}{c|}{Automatic($\uparrow$)}
& \multicolumn{2}{c}{Unfaithfulness(\%$\downarrow$)} & \multicolumn{2}{c}{Automatic($\uparrow$)} \\
& & Counter & Total & SacreBLEU & BERTScore & Counter & Total & SacreBLEU & BERTScore \\
\midrule
\multicolumn{10}{c}{\textbf{T5-based}} \\ 
\midrule
- & \textbf{\texttt{Fine-tuning\textsubscript{base}}} &
  47.08 & 16.89 & 15.430 & 0.894 &
  \textbf{87.17} &73.73  & 7.634 & 0.876 \\
\midrule
\multirow{3}{*}{Highlight Token} & \textbf{\texttt{Prompt}} &
  42.04 & 14.11 & 15.686 & 0.898 &
  87.04 & 74.18 & 7.592 & 0.876 \\
& \texttt{\textbf{\gat}} (Ours) &
  35.92 & 11.28 & 16.106 & \textbf{0.899} &
  91.75 & 57.51 & \textbf{8.990} & 0.883 \\
& \texttt{\textbf{\gcn}} (Ours) &
  35.47 & 10.88 & 16.111 & \textbf{0.899} &
  92.13 & \textbf{57.00} & 8.672 & 0.881 \\
\midrule
\multirow{3}{*}{Token Interactions} & \textbf{\texttt{Prompt}} &
  51.56 & 19.2 & 15.478 & 0.898 &
  87.49 & 76.43 & 7.028 & 0.888  \\
& \texttt{\textbf{\gat}} (Ours) &
  \textbf{34.28} & 10.67 & 16.106 & \textbf{0.899} &
  92.04 & 74.60 & 7.692 & \textbf{0.891}  \\
& \texttt{\textbf{\gcn}} (Ours) &
  32.59 & \textbf{10.03} & 16.121 & \textbf{0.899} &
  92.75 & 77.03 & 7.831 & \textbf{0.891}  \\
\midrule
\multirow{3}{*}{Span Interactions} & \textbf{\texttt{Prompt}} &
  42.47 & 13.65 & 16.031 & \textbf{0.899} &
  89.34 & 79.44 & 7.226 & 0.815  \\
& \texttt{\textbf{\gat}} (Ours) &
  38.05 & 12.05 & 16.119 & \textbf{0.899} &
  92.73 & 68.15 & 7.256 & 0.815  \\
& \texttt{\textbf{\gcn}} (Ours) &
  34.31 & 10.82 & \textbf{16.160} & 0.898 &
  91.99 & 71.77 & 7.771 & \textbf{0.891}  \\
\midrule[1.5pt]
\multicolumn{10}{c}{\textbf{BART-based}} \\ 
\midrule
- & \textbf{\texttt{Fine-tuning\textsubscript{base}}} &
  57.98 & 19.64 & 15.732 & 0.906 &
  \textbf{82.72} & 72.82 & 10.070 & \textbf{0.891} \\
\midrule
\multirow{3}{*}{Highlight Token} & \textbf{\texttt{Prompt}} &
  56.65 & 24.20 & 15.678 & 0.898 &
  84.74 & 61.97 & 10.012 & \textbf{0.891} \\
& \texttt{\textbf{\gat}} (Ours) &
  43.85 & 13.78 & \textbf{16.503} & \textbf{0.909} &
  91.97 & \textbf{58.11} & 10.092 & \textbf{0.891} \\
& \texttt{\textbf{\gcn}} (Ours) &
  44.68 & 14.32  &16.364 & \textbf{0.909} &
  90.95 & 59.13 & \textbf{10.489} & 0.893  \\
\midrule
\multirow{3}{*}{Token Interactions} & \textbf{\texttt{Prompt}} &
  51.56 & 19.20 & 15.478 & 0.898 &
  95.85 & 69.86 & 7.868 & 0.890 \\
& \texttt{\textbf{\gat}} (Ours) &
  48.38 & 16.07 & 16.24 & 0.908 &
  95.21 & 72.52 & 7.405 & 0.888 \\
& \texttt{\textbf{\gcn}} (Ours) &
  \textbf{41.57} & \textbf{12.89} & 16.364 & \textbf{0.909} &
  94.11 & 72.03 & 7.700 & 0.889 \\
\midrule
\multirow{3}{*}{Span Interactions} & \textbf{\texttt{Prompt}} &
  51.10 & 17.41 & 16.046 & 0.888 &
  94.89 & 65.52 & 7.333 & 0.888 \\
& \texttt{\textbf{\gat}} (Ours) &
  42.90 & 12.92 & 16.449 & \textbf{0.909} &
  93.98 & 61.39 & 7.795 & 0.890 \\
& \texttt{\textbf{\gcn}} (Ours) &
  45.48 & 14.10 & 16.447 & \textbf{0.909} &
  71.07 & 96.44 & 7.518 & 0.887 \\
\bottomrule
\end{tabular}
}
\caption{Overall evaluation results on e-SNLI and ComVE datasets for T5-based and BART-based models, with our \texttt{\textbf{\method}} model using \texttt{\textbf{\gat}} and \texttt{\textbf{\gcn}}. 
Counter indicates \textit{Counter Unfaith}, Total indicates \textit{Total Unfaith}. The best performance of each evaluation metric is in bold. See Table \ref{tab:esnli_comve_results_t5_and_bart} for results of our model using \texttt{\textbf{\sage}}.}
\label{tab:esnli_comve_results_t5_and_bart_gcn_and_gat}
\end{table*}

\section{Statistical Uncertainty Measurement for Faithfulness Evaluation on e-SNLI and ComVE Datasets using \texttt{\textbf{\sage}} and \textbf{\texttt{Fine-tuning\textsubscript{base}}} with \texttt{T5-large} and \texttt{BART-large} models}
\label{sec:appendix:esnli_comve_uncertainty}

To demonstrate the significant improvement of our \texttt{\textbf{\method}} in terms of faithfulness, we compute the p-values ~\cite{wasserstein2016asa} for \textit{Counter Unfaith} and \textit{Total Unfaith} (see Section \S\ref{sec:results:Faithfulness Evaluation}) when comparing the \textbf{\texttt{Fine-tuning\textsubscript{base}}} and our \texttt{\textbf{\sage}} model on the e-SNLI and ComVE datasets, using \texttt{T5-large} and \texttt{BART-large} with 5 random seeds.
\begin{table}[ht] 
\centering
\scalebox{0.65}{
\begin{tabular}{@{}c c|c|c|c|c@{}}
\toprule
\multirow{2}{*}{\textbf{Explanation Type}} & \multirow{2}{*}{\textbf{Model}}  
& \multicolumn{2}{c|}{\textbf{e-SNLI (P-Value)}} & \multicolumn{2}{c}{\textbf{ComVE (P-Value)}} \\
\cmidrule(lr){3-4}\cmidrule(lr){5-6}
& & Counter & Total & Counter & Total \\
& & Unfaith & Unfaith & Unfaith & Unfaith \\
\midrule
\multicolumn{6}{c}{\textbf{T5-based}} \\
\midrule
Highlight Token & \texttt{\textbf{\sage}} & 0.0007 & 0.0054 & 0.0136 & 0.0002 \\
Token Interactions & \texttt{\textbf{\sage}} & 0.0002 & 0.0001 & 0.0164 & 0.0047 \\
Span Interactions & \texttt{\textbf{\sage}} & 0.0010 & 0.0032 & 0.0001 & 0.0307 \\
\midrule
\multicolumn{6}{c}{\textbf{BART-based}} \\
\midrule
Highlight Token & \texttt{\textbf{\sage}} & 0.0067 & 0.0064 & 0.0455 & 0.0001 \\
Token Interactions & \texttt{\textbf{\sage}} & 0.0122 & 0.0007 & 0.0168 & 0.0169 \\
Span Interactions & \texttt{\textbf{\sage}} & 0.0033 & 0.0006 & 0.0403 & 0.0116 \\
\bottomrule
\end{tabular}}
\caption{P-values of our \texttt{\textbf{\sage}} model compared to \textbf{\texttt{Fine-tuning\textsubscript{base}}} on the e-SNLI and ComVE datasets, using \texttt{T5-large} and \texttt{BART-large}, regarding \textit{Counter Unfaith} and \textit{Total Unfaith} on 5 random seeds. }
\label{tab:uncertainty}
\end{table}

As shown in Table \ref{tab:uncertainty}, all p-values are less than 0.05, indicating that the natural language explanations generated by our \texttt{\textbf{\method}} exhibit significantly lower unfaithfulness compared to the baseline method.

\section{Overall Explanation Evaluation Results on ECQA dataset for \method based on \texttt{T5-large} and \texttt{BART-large}}
\label{sec:appendix:ecqa_t5_bart}

\subsection{Overall Explanation Evaluation Results on ECQA dataset for \method based on \texttt{T5-large} }

The faithfulness and automatic evaluation results of T5-based models on the ECQA dataset are shown in Table \ref{tab:all_results_ecqa_t5}. 

Regarding the faithfulness of NLEs, almost all of our methods outperform the baseline methods, highlighting the effectiveness of our framework. Among the different highlight explanation types, \textbf{\textit{token interactive explanations}} demonstrate the best performance in generating faithful NLEs when using \textbf{\textit{\gcn}}, achieving 21.18\% total unfaithfulness. Other variants, such as \textbf{\textit{\gat}} and \textbf{\textit{\sage}}, also achieve comparable performance, with 21.44\% and 21.74\% total unfaithfulness, respectively. \textbf{On the ECQA dataset, \textbf{\textit{token interactive explanations}} show a clear advantage over other highlight explanation types in improving the faithfulness of NLEs}.

Regarding the similarity between the generated NLEs and the gold ones, \method outperforms the fine-tuning baseline in most settings. Although the prompt baseline achieves the highest SacreBLEU and BERTScore, \method lags behind by only 1.537 in SacreBLEU and 0.004 in BERTScore. Among all types of highlight explanations, \textbf{\textit{span interactive explanations}} achieve the highest scores with \method.

\begin{table*}[ht]
\centering
\small
\scalebox{1.0}{
\begin{tabular}{cccccc}
\toprule
\multicolumn{2}{c}{\multirow{2}{*}{\makecell{Evaluation \\ Metrics}}} & \multicolumn{2}{c}{UnFaithfulness(\% $\downarrow$)} & \multicolumn{2}{c}{Automatic Evaluation ($\uparrow$)} \\
\cmidrule(lr){3-4}\cmidrule(lr){5-6}
& & \makecell{Counter \\Unfaith} & \makecell{Total \\Unfaith} & \makecell{SacreBLEU \\ (0-100)} & \makecell{BERTScore \\ (0-1)} \\
\midrule
\multicolumn{2}{c}{\textbf{\texttt{Fine-tuning\textsubscript{base}}}} & 49.34 & 24.80 & 14.057 & 0.883 \\
\midrule
\multirow{4}{*}{\makecell{Highlight Tokens}} 
& \textbf{\texttt{Prompt}} & 46.56 & 25.27 & 15.303 & 0.887 \\
& \texttt{\textbf{\gat}} & 44.76 & 21.99 & 14.048 & 0.883 \\
& \texttt{\textbf{\gcn}} & 49.61 & 25.21 & 13.855 & 0.882 \\
& \texttt{\textbf{\sage}} & 45.42 & 22.44 & 13.968 & 0.882 \\
\midrule
\multirow{4}{*}{\makecell{Token Interactions}}
& \textbf{\texttt{Prompt}} & 51.29 & 33.30 & 15.311 & 0.887 \\
& \texttt{\textbf{\gat}} & 43.49 & 21.44 & 13.910 & 0.882 \\
& \texttt{\textbf{\gcn}} & \textbf{43.42} & \textbf{21.18} & 14.079 & 0.883 \\
& \texttt{\textbf{\sage}} & 44.20 & 21.74 & 13.978 & 0.882 \\
\midrule
\multirow{4}{*}{\makecell{Span Interactions}}
& \textbf{\texttt{Prompt}} & 50.20 & 28.22 & \textbf{16.046} & \textbf{0.888} \\
& \texttt{\textbf{\gat}} & 49.22 & 23.85 & 14.339 & 0.883 \\
& \texttt{\textbf{\gcn}} & 50.46 & 24.91 & 14.477 & 0.883 \\
& \texttt{\textbf{\sage}} & 46.87 & 22.50 & 14.509 & 0.884 \\
\bottomrule
\end{tabular}
}
\caption{Overall Evaluation Results on ECQA of T5-based \method. The best performance of each evaluation metric across all NLE generation models is in bold.}
\label{tab:all_results_ecqa_t5}
\end{table*}

\subsection{Automatic Evaluation Results on ECQA dataset for \method based on \texttt{BART-large} }
As shown in Table \ref{tab:automatic_results_ecqa_bart}, we also conduct automatic evaluation on BART-based \method on ECQA datasets regarding Lexical and Semantical Similarity with golden explanations. 

Compared to all the baseline methods, on ECQA dataset, with the highest scores always belong to our \textit{token interactive explanation} guided \texttt{\textbf{\gcn}} method, and other variants are with comparable performance to the baselines, our model also shows advantage in both lexical and semantic similarity. 

Among the different explanation types, \textit{token interactive explanations} demonstrate superior performance in both lexical and semantic metrics. Notably, \textbf{\textit{token interactive explanations} show a slight advantage over the other two explanation types in generating NLEs with more plausible meanings to humans.}

\begin{table*}[ht]
\centering
\small
\normalsize
\scalebox{0.6}{
\begin{tabular}{cc c c c c c c}
\toprule
\multicolumn{2}{c}{\multirow{2}{*}{Automatic Evaluation Metrics}} & \multicolumn{3}{c}{Lexical Similarity ($\uparrow$)} & \multicolumn{3}{c}{Semantic Similarity ($\uparrow$)} \\
\cmidrule(lr){3-5}\cmidrule(lr){6-8}
\multicolumn{2}{c}{} & ROUGE-1 (0-1) & ROUGE-L (0-1) & SacreBLEU (1-100) & MoverScore (0-1) & BARTScore (-0-1) & BERTScore (0-1) \\
\midrule
\multicolumn{2}{c}{\textbf{\texttt{Fine-tuning\textsubscript{base}}}} & 0.180 & 0.130 & 12.484 & 0.840 & -4.433 & 0.836 \\
\midrule
\multicolumn{1}{c}{\multirow{4}{*}{Highlight Tokens}} & \textbf{\texttt{Prompt}} & 0.112 & 0.077 & 10.733 & 0.767 & -4.557 & 0.754 \\
& \texttt{\textbf{\gat}} (Ours) & 0.172 & 0.125 & 12.186 & 0.837 & -4.453 & 0.835 \\
& \texttt{\textbf{\gcn}} (Ours) & 0.198 & 0.146 & 13.091 & 0.840 & -4.379 & 0.839 \\
& \texttt{\textbf{\sage}} (Ours) & 0.181 & 0.133 & 12.659 & 0.839 & -4.434 & 0.836 \\
\midrule
\multicolumn{1}{c}{\multirow{4}{*}{Token Interactions}} & \textbf{\texttt{Prompt}} & 0.185 & 0.134 & 12.724 & 0.838 & -4.435 & 0.837 \\
& \texttt{\textbf{\gat}} (Ours) & 0.208 & 0.151 & 13.519 & 0.841 & -4.399 & 0.841 \\
& \texttt{\textbf{\gcn}} (Ours) & \textbf{0.321} & \textbf{0.226} & \textbf{17.860} & \textbf{0.848} & \textbf{-4.079} & \textbf{0.858} \\
& \texttt{\textbf{\sage}} (Ours) & 0.243 & 0.174 & 14.773 & 0.843 & -4.269 & 0.847 \\
\midrule
\multicolumn{1}{c}{\multirow{4}{*}{pan Interactions}} & \textbf{\texttt{Prompt}} & 0.175 & 0.126 & 12.288 & 0.839 & -4.454 & 0.835 \\
& \texttt{\textbf{\gat}} (Ours) & 0.176 & 0.128 & 12.295 & 0.838 & -4.456 & 0.835 \\
& \texttt{\textbf{\gcn}} (Ours) & 0.175 & 0.128 & 12.364 & 0.838 & -4.455 & 0.835 \\
& \texttt{\textbf{\sage}} (Ours) & 0.186 & 0.135 & 12.802 & 0.839 & -4.415 & 0.837 \\
\bottomrule
\end{tabular}
}
\caption{Automatic Evaluation Results on ECQA of BART-based \method. The best performance of each evaluation metric across different NLE generation models is in bold.}
\label{tab:automatic_results_ecqa_bart}
\end{table*}

\subsection{Faithfulness Evaluation Results on ECQA dataset for \method based on \texttt{BART-large} }
We also evaluated the faithfulness of \method based on \texttt{BART-large} on the ECQA dataset and observed that the faithfulness scores for all methods (including the baselines) were uniformly 100\%. This result indicates that the BART-based models are prone to counterfactual attacks and none of these explanations were faithful. We attribute this outcome to the inherent complexity of the ECQA dataset and the potential vulnerability of the BART model to counterfactual attacks.

\section{Supplementary Automatic Explanation Evaluation Results for \method based on \texttt{T5-large} and \texttt{BART-large}}
\label{sec:appendix:t5_other_automatic_metrics}

To evaluate the similarity between the generated NLE and the golden ones as an approximation of plausibility to humans, we also leverage the following four metrics to evaluate their lexical and semantic similarity:

\textbf{Rouge1}~\citep{lin-2004-rouge} calculates the overlap of unigrams between the generated explanation and the golden ones, providing insight into lexical similarity at the word level.

\textbf{RougeL}~\citep{lin-2004-rouge} measures the longest common subsequence between the generated explanation and the golden explanations.

\textbf{MoverScore}~\citep{zhao-etal-2019-moverscore} calculates semantic similarity by computing word embeddings and their movement cost, capturing meaning while accounting for variations in word order and structure.

\textbf{BARTScore}~\citep{NEURIPS2021_e4d2b6e6} leverages BART's language model to assess the likelihood of the reference text being generated given the generated explanation as input, providing a fluency and relevance measure.

\subsection{Supplementary Automatic Explanation Evaluation Results for \method based on \texttt{T5-large}}

As shown in Table \ref{tab:part_automatic_results_esnli_t5}, Table \ref{tab:part_of_automatic_results_ecqa_t5} and Table \ref{tab:part_of_automatic_results_comve_t5}, we conduct a supplementary automatic evaluation on T5-based \method regarding Lexical Similarity and Semantic Similarity with the golden explanations on e-SNLI, ECQA and ComVE datasets respectively. 

Compared to all the baseline methods on the e-SNLI dataset, \textbf{all variants of our \method achieve higher lexical and semantic similarity with gold explanations}, indicating that our approach can generate more plausible NLEs. For instance, we observe up to a 2.1\% improvement in ROUGE-1 and a notable absolute increase of 0.224 in BARTScore. On the ECQA dataset, our \method achieves better similarity performance than \texttt{Fine-tuning\textsubscript{base}} (which does not utilize explanation information) and is comparable to the prompt-based baseline. On the ComVE dataset, all NLEs generated by our method incorporating \textbf{\textit{highlight token explanations}} surpass the baselines in both lexical and semantic similarity, while the variants based on \textbf{\textit{token interactive explanations}} and \textbf{\textit{span interactive explanations}} sometimes fail to do so. This is likely due to the format of the ComVE dataset, which presents a simple question followed by two similar statements. In this scenario, \textbf{\textit{token interactive explanations}} and \textbf{\textit{span interactive explanations}} may struggle to capture sufficient information from the limited interaction between the question and the options.

Among the different highlight explanation types on the e-SNLI dataset, \textbf{\textit{token interactive explanations}}, particularly those using the \sage variant of our \method, achieve the highest lexical and semantic similarity. Meanwhile, \textbf{\textit{highlight token explanations}} and \textbf{\textit{span interactive explanations}} also perform strongly, excelling at ROUGE-L and ROUGE-1 scores respectively. On the ECQA dataset, \textbf{\textit{span interactive explanations}} have a slight edge over other explanation types, although the difference is marginal. On the ComVE dataset, \textbf{\textit{highlight token explanations}} show a clear advantage across all metrics. This is likely due to the input format of the ComVE dataset, which makes it challenging for \textbf{\textit{token interactive explanations}} and \textbf{\textit{span interactive explanations}} to capture sufficient information, as discussed earlier.

In summary, \textbf{these findings highlight that the advantages of different explanation types in improving NLE quality vary with dataset characteristics.}

\begin{table*}[ht]
\centering
\small
\normalsize
\scalebox{0.8}{
\begin{tabular}{cc c c c c}
\toprule
\multicolumn{2}{c}{\multirow{2}{*}{Automatic Evaluation Metrics}} & \multicolumn{2}{c}{Lexical Similarity($\uparrow$)} & \multicolumn{2}{c}{Semantic Similarity($\uparrow$)} \\
\cmidrule(lr){3-4}\cmidrule(lr){5-6}
\multicolumn{2}{c}{} & ROUGE-1 (0-1) & ROUGE-L (0-1) & MoverScore (0-1) & BARTScore (-0-1) \\
\midrule
\multicolumn{2}{c}{\textbf{\texttt{Fine-tuning\textsubscript{base}}}} & 0.448 & 0.384 & 0.838 & -3.646 \\
\midrule
\multicolumn{1}{c}{\multirow{4}{*}{Highlight Tokens}} & \textbf{\texttt{Prompt}} & 0.455 & 0.397 & 0.840 & -3.492 \\
& \texttt{\textbf{\gat}} (Ours) & 0.467 & 0.402 & 0.842 & -3.437 \\
& \texttt{\textbf{\gcn}} (Ours) & 0.468 & 0.403 & 0.842 & -3.425 \\
& \texttt{\textbf{\sage}} (Ours) & 0.468 & \textbf{0.404} & 0.841 & \textbf{-3.422} \\
\midrule
\multicolumn{1}{c}{\multirow{4}{*}{Token Interactions}} & \textbf{\texttt{Prompt}} & 0.459 & 0.394 & 0.842 & -3.503 \\
& \texttt{\gat} (Ours) & 0.467 & 0.402 & 0.842 & -3.437 \\
& \texttt{\textbf{\gcn}} (Ours) & 0.467 & 0.403 & 0.842 & -3.435 \\
& \texttt{\textbf{\sage}} (Ours) & \textbf{0.469} & \textbf{0.404} & \textbf{0.843} & -3.431 \\
\midrule
\multicolumn{1}{c}{\multirow{4}{*}{Span Interactions}} & \textbf{\texttt{Prompt}} & 0.466 & 0.402 & 0.841 & -3.467 \\
& \texttt{\gat} (Ours) & 0.466 & 0.403 & 0.841 & -3.442 \\
& \texttt{\textbf{\gcn}} (Ours) & \textbf{0.469} & 0.403 & 0.843 & -3.433 \\
& \texttt{\textbf{\sage}} (Ours) & 0.467 & 0.402 & 0.842 & -3.428 \\
\bottomrule
\end{tabular}
}
\caption{Automatic Evaluation Results on e-SNLI of T5-based \method (excluding SacreBLEU and BERTScore, which are presented in Table \ref{tab:esnli_comve_results_t5_and_bart}). The best performance of each evaluation metric across different NLE generation models is in bold.}
\label{tab:part_automatic_results_esnli_t5}
\end{table*}

\begin{table*}[ht]
\centering
\small
\normalsize
\scalebox{0.8}{
\begin{tabular}{cc c c c c}
\toprule
\multicolumn{2}{c}{\multirow{2}{*}{Automatic Evaluation Metrics}} & \multicolumn{2}{c}{Lexical Similarity($\uparrow$)} & \multicolumn{2}{c}{Semantic Similarity($\uparrow$)} \\ 
\cmidrule(lr){3-4}\cmidrule(lr){5-6}
\multicolumn{2}{c}{} & ROUGE-1 (0-1) & ROUGE-L (0-1) & MoverScore (0-1) & BARTScore (-0-1) \\ 
\midrule
\multicolumn{2}{c}{\textbf{\texttt{Fine-tuning\textsubscript{base}}}} & 0.469 & 0.346 & 0.850 & -3.584 \\
\midrule
\multicolumn{1}{c}{\multirow{4}{*}{Highlight Tokens}} & \textbf{\texttt{Prompt}} & 0.490 & 0.355 & \textbf{0.857} & \textbf{-3.528} \\
& \texttt{\textbf{\gat}} (Ours) & 0.469 & 0.346 & 0.851 & -3.576 \\
& \texttt{\textbf{\gcn}} (Ours) & 0.468 & 0.347 & 0.850 & -3.575 \\
& \texttt{\textbf{\sage}} (Ours) & 0.468 & 0.347 & 0.850 & -3.569 \\
\midrule
\multicolumn{1}{c}{\multirow{4}{*}{Token Interactions}} & \textbf{\texttt{Prompt}} & 0.489 & 0.354 & 0.855 & -3.549 \\
& \texttt{\textbf{\gat}} (Ours) & 0.468 & 0.345 & 0.849 & -3.598 \\
& \texttt{\textbf{\gcn}} (Ours) & 0.469 & 0.346 & 0.850 & -3.593 \\
& \texttt{\textbf{\sage}} (Ours) & 0.468 & 0.346 & 0.851 & -3.593 \\
\midrule
\multicolumn{1}{c}{\multirow{4}{*}{Span Interactions}} & \textbf{\texttt{Prompt}} & \textbf{0.496} & \textbf{0.360} & \textbf{0.857} & -3.520 \\
& \texttt{\textbf{\gat}} (Ours) & 0.472 & 0.350 & 0.850 & -3.569 \\
& \texttt{\textbf{\gcn}} (Ours) & 0.470 & 0.349 & 0.849 & -3.568 \\
& \texttt{\textbf{\sage}} (Ours) & 0.474 & 0.350 & 0.851 & -3.560 \\
\bottomrule
\end{tabular}
}
\caption{Automatic Evaluation Results on ECQA of T5-based \method (excluding SacreBLEU and BERTScore, which are presented in Table \ref{tab:all_results_ecqa_t5}). The best performance of each evaluation metric across different NLE generation model is in bold.}
\label{tab:part_of_automatic_results_ecqa_t5}
\end{table*}

\begin{table*}[ht]
\centering
\small
\normalsize
\scalebox{0.8}{
\begin{tabular}{cc c c c c}
\toprule
\multicolumn{2}{c}{\multirow{2}{*}{Automatic Evaluation Metrics}} & \multicolumn{2}{c}{Lexical Similarity($\uparrow$)} & \multicolumn{2}{c}{Semantic Similarity($\uparrow$)} \\ 
\cmidrule(lr){3-4}\cmidrule(lr){5-6}
\multicolumn{2}{c}{} & ROUGE-1 (0-1) & ROUGE-L (0-1) & MoverScore (0-1) & BARTScore (-0-1) \\ 
\midrule
\multicolumn{2}{c}{\textbf{\texttt{Fine-tuning\textsubscript{base}}}} & 0.355 & 0.319 & 0.828 & -4.030 \\
\midrule
\multicolumn{1}{c}{\multirow{4}{*}{Highlight Tokens}} & \textbf{\texttt{Prompt}} & 0.354 & 0.317 & 0.825 & -4.051 \\
& \texttt{\textbf{\gat}} (Ours) & \textbf{0.394} & 0.332 & 0.832 & -3.884 \\
& \texttt{\textbf{\gcn}} (Ours) & 0.384 & \textbf{0.333} & 0.830 & -3.934 \\
& \texttt{\textbf{\sage}} (Ours) & 0.393 & 0.330 & \textbf{0.833} & \textbf{-3.881} \\
\midrule
\multicolumn{1}{c}{\multirow{4}{*}{Token Interactions}} & \textbf{\texttt{Prompt}} & 0.312 & 0.269 & 0.817 & -4.083 \\
& \texttt{\textbf{\gat}} (Ours) & 0.326 & 0.283 & 0.816 & -3.976 \\
& \texttt{\textbf{\gcn}} (Ours) & 0.332 & 0.288 & 0.817 & -3.970 \\
& \texttt{\textbf{\sage}} (Ours) & 0.310 & 0.266 & 0.817 & -4.070 \\
\midrule
\multicolumn{1}{c}{\multirow{4}{*}{Span Interactions}} & \textbf{\texttt{Prompt}} & 0.317 & 0.275 & 0.815 & -4.059 \\
& \texttt{\textbf{\gat}} (Ours) & 0.324 & 0.280 & 0.815 & -3.998 \\
& \texttt{\textbf{\gcn}} (Ours) & 0.328 & 0.286 & 0.815 & -3.975 \\
& \texttt{\textbf{\sage}} (Ours) & 0.328 & 0.283 & 0.818 & -3.980 \\
\bottomrule
\end{tabular}
}
\caption{Automatic Evaluation Results on ComVE of T5-based \method (excluding SacreBLEU and BERTScore, which are presented in Table \ref{tab:esnli_comve_results_t5_and_bart}). The best performance of each evaluation metric across different NLE generation models is in bold.}
\label{tab:part_of_automatic_results_comve_t5}
\end{table*}

\subsection{Supplementary Automatic Explanation Evaluation Results for \method based on \texttt{BART-large}}

As shown in Table \ref{tab:part_automatic_results_esnli_bart}, Table \ref{tab:automatic_results_ecqa_bart} and Table \ref{tab:part_automatic_results_comve_bart}, we conduct a supplementary automatic evaluation on BART-based \method regarding Lexical Similarity and Semantic Similarity with the golden explanations on e-SNLI, ECQA and ComVE datasets respectively.


\begin{table*}[ht]
\centering
\small
\normalsize
\scalebox{0.8}{
\begin{tabular}{cc c c c c}
\toprule
\multicolumn{2}{c}{\multirow{2}{*}{Automatic Evaluation Metrics}} & \multicolumn{2}{c}{Lexical Similarity($\uparrow$)} & \multicolumn{2}{c}{Semantic Similarity($\uparrow$)} \\
\cmidrule(lr){3-4}\cmidrule(lr){5-6}
\multicolumn{2}{c}{} & ROUGE-1 (0-1) & ROUGE-L (0-1) & MoverScore (0-1) & BARTScore (-0-1) \\
\midrule
\multicolumn{2}{c}{\textbf{\texttt{Fine-tuning\textsubscript{base}}}} & 0.457 & 0.391 & 0.838 & -3.491 \\
\midrule
\multicolumn{1}{c}{\multirow{4}{*}{Highlight Tokens}} & \textbf{\texttt{Prompt}} & 0.468 & 0.398 & \textbf{0.843} & -3.458 \\
& \texttt{\textbf{\gat}} (Ours) & 0.476 & 0.405 & \textbf{0.843} & \textbf{-3.403} \\
& \texttt{\textbf{\gcn}} (Ours) & 0.474 & 0.402 & 0.841 & -3.415 \\
& \texttt{\textbf{\sage}} (Ours) & 0.474 & 0.402 & 0.840 & -3.416 \\
\midrule
\multicolumn{1}{c}{\multirow{4}{*}{Token Interactions}} & \textbf{\texttt{Prompt}} & 0.459 & 0.394 & \textbf{0.843} & -3.503 \\
& \texttt{\textbf{\gat}} (Ours) & 0.472 & 0.401 & 0.841 & -3.449 \\
& \texttt{\textbf{\gcn}} (Ours) & 0.473 & 0.402 & 0.842 & -3.418 \\
& \texttt{\textbf{\sage}} (Ours) & 0.472 & 0.403 & 0.841 & -3.431 \\
\midrule
\multicolumn{1}{c}{\multirow{4}{*}{Span Interactions}} & \textbf{\texttt{Prompt}} & 0.475 & 0.403 & 0.841 & -3.419 \\
& \texttt{\textbf{\gat}} (Ours) & \textbf{0.477} & 0.403 & 0.842 & -3.427 \\
& \texttt{\textbf{\gcn}} (Ours) & 0.476 & 0.403 & 0.842 & -3.423 \\
& \texttt{\textbf{\sage}} (Ours) & \textbf{0.477} & \textbf{0.404} & 0.842 & -3.423 \\
\bottomrule
\end{tabular}
}
\caption{Automatic Evaluation Results on e-SNLI of BART-based \method (SacreBLEU and BERTScore are excluded and are presented in Table \ref{tab:esnli_comve_results_t5_and_bart}). The best performance of each evaluation metric across different NLE generation models is in bold.}
\label{tab:part_automatic_results_esnli_bart}
\end{table*}

\begin{table*}[ht]
\centering
\small
\normalsize
\scalebox{0.8}{
\begin{tabular}{cc c c c c}
\toprule
\multicolumn{2}{c}{\multirow{2}{*}{Automatic Evaluation Metrics}} & \multicolumn{2}{c}{Lexical Similarity($\uparrow$)} & \multicolumn{2}{c}{Semantic Similarity($\uparrow$)} \\ 
\cmidrule(lr){3-4}\cmidrule(lr){5-6}
\multicolumn{2}{c}{} & ROUGE-1 (0-1) & ROUGE-L (0-1) & MoverScore (0-1) & BARTScore (-0-1) \\ 
\midrule
\multicolumn{2}{c}{\textbf{\texttt{Fine-tuning\textsubscript{base}}}} & 0.421 & 0.325 & \textbf{0.840} & -3.802 \\
\midrule
\multicolumn{1}{c}{\multirow{4}{*}{Highlight Tokens}} & \textbf{\texttt{Prompt}} & 0.419 & 0.322 & 0.834 & -3.796 \\
& \texttt{\textbf{\gat}} (Ours) & 0.427 & 0.325 & 0.837 & -3.765 \\
& \texttt{\textbf{\gcn}} (Ours) & \textbf{0.435} & \textbf{0.332} & 0.838 & -3.761 \\
& \texttt{\textbf{\sage}} (Ours) & 0.434 & 0.330 & 0.837 & \textbf{-3.748} \\
\midrule
\multicolumn{1}{c}{\multirow{4}{*}{Token Interactions}} & \textbf{\texttt{Prompt}} & 0.334 & 0.284 & 0.818 & -4.036 \\
& \texttt{\textbf{\gat}} (Ours) & 0.322 & 0.277 & 0.818 & -4.047 \\
& \texttt{\textbf{\gcn}} (Ours) & 0.334 & 0.285 & 0.817 & -3.985 \\
& \texttt{\textbf{\sage}} (Ours) & 0.316 & 0.269 & 0.814 & -4.129 \\
\midrule
\multicolumn{1}{c}{\multirow{4}{*}{Span Interactions}} & \textbf{\texttt{Prompt}} & 0.323 & 0.274 & 0.818 & -4.029 \\
& \texttt{\textbf{\gat}} (Ours) & 0.334 & 0.288 & 0.818 & -4.011 \\
& \texttt{\textbf{\gcn}} (Ours) & 0.327 & 0.278 & 0.818 & -4.045 \\
& \texttt{\textbf{\sage}} (Ours) & 0.333 & 0.287 & 0.820 & -4.017 \\
\bottomrule
\end{tabular}
}
\caption{Automatic Evaluation Results on ComVE of BART-based \method. The best performance of each evaluation metric across different NLE generation models is in bold.}
\label{tab:part_automatic_results_comve_bart}
\end{table*}

\section{Human Evaluation}
\label{Human Evaluation}
In line with prior work \citep{atanasova-etal-2020-generating-fact,jolly2022generating}, our human evaluation assesses the generated explanations across four key dimensions: 

\textbf{Coverage}: The explanation includes all important and salient information, ensuring no significant points that contribute to label prediction are omitted.

\textbf{Non-redundancy}: The explanation should avoid redundant, repeated, or irrelevant information and should not include content that is unreasonable or inconsistent with common sense.

\textbf{Non-contradiction}: The explanation should not contradict the predicted label or the input text, maintaining consistency throughout.

\textbf{Overall Quality}: The explanations are rated based on overall quality, considering factors such as grammar, readability, and clarity.

We engaged three PhD students with backgrounds in computer science to evaluate the explanations using a 1–7 Likert scale following previous work \citep{castro-ferreira-etal-2019-neural,ribeiro-etal-2021-investigating,yuan-farber-2024-grasame}. We compare the text generated by the \texttt{Fine-tuning\textsubscript{base}} with that generated by \texttt{\textbf{\gat}} when guided by \textit{highlight token}, \textit{token interactive explanations}, and \textit{span interactive explanations}, respectively.  
The annotator agreement is reported in Table \ref{tab:agreemeent}. Note that we randomly sample 100 NLEs generated by each model.  


\subsection{Human Evaluation Results}
\begin{table*}[ht]
\centering
\normalsize
\begin{tabular}{ccccc}
\toprule
\textbf{Method} & \textbf{Coverage} & \textbf{Non Redund.} & \textbf{Non Contrad.} & \textbf{Overall} \\
\midrule
\texttt{Fine-tuning\textsubscript{base}} & 6.72 & 5.86 & 6.67 & 6.28 \\
Highlight Tokens & 6.74 & 5.80 & 6.67 & 6.06 \\
Token Interactions & \textbf{6.75} & \textbf{5.95} & 6.64 & \textbf{6.37} \\
Span Interactions & 6.67 & 5.92 & \textbf{6.72} & 6.26 \\
\bottomrule
\end{tabular}
\caption{Human Evaluation Results on e-SNLI dataset of our \texttt{\textbf{\method}} using \texttt{\textbf{\gat}} based on T5.}
\label{tab:esnli_human_eval}
\end{table*}

\begin{table*}[ht]
\centering
\normalsize
\begin{tabular}{ccccc}
\toprule
\textbf{Method} & \textbf{Coverage} & \textbf{Non Redund.} & \textbf{Non Contrad.} & \textbf{Overall} \\
\midrule
\texttt{Fine-tuning\textsubscript{base}} & 5.66 & 4.41 & 4.91 & 5.53 \\
Highlight Tokens  & 5.08 & 4.27 & 4.51 & 5.20 \\
Token Interactions & 5.60 & \textbf{4.82} & \textbf{5.08} & 5.61 \\
Span Interactions  & \textbf{5.65} & 4.67 & 4.90 & \textbf{5.63} \\
\bottomrule
\end{tabular}
\caption{Human Evaluation Results on ECQA dataset of  our \texttt{\textbf{\method}} using \texttt{\textbf{\gat}} based on T5.}
\label{tab:ecqa_human_eval}
\end{table*}

\paragraph{e-SNLI} In Table \ref{tab:esnli_human_eval}, across all highlight explanation types, the NLEs generated by the \textit{token interactive explanations} achieve the highest scores across most dimensions, particularly excelling in \textit{Non-redundancy} (5.95) and \textit{Overall Quality} (6.37), indicating its effectiveness in producing concise and high-quality explanations. The NLEs generated with the guidance of \textit{span interactive explanations} method also show strong performance, especially in \textit{Non-contradiction} (6.72), suggesting that modeling span-level interactions is beneficial for maintaining consistency of the NLE with the generated label. The \textit{highlighted token explanations} performs slightly lower, indicating that while it captures key tokens effectively, it may miss out on broader contextual relationships crucial for non-redundancy and overall quality.

\paragraph{ECQA} Table \ref{tab:ecqa_human_eval} shows the evaluation results for the ECQA dataset, where the NLEs generated by \textit{token interactive explanations} again lead in \textit{Non-redundancy} (4.82) and achieves a high \textit{Non-contradiction} score (5.08), confirming its robustness across different datasets. The \textit{span interactive explanations} perform similarly well, attaining the highest \textit{Overall Quality} score (5.63), emphasizing its adaptability in varied datasets. 

Overall, while the \textit{highlight token explanations} shows slightly lower performance across all highlight explanation types, leveraging \textit{span interactive explanations} and \textit{token interactive explanations} that are encoded in \method notably improves the quality and consistency of the generated explanations.

\subsection{Human Evaluation Instruction\label{human_instruction}}

The annotators are asked to rate the generated texts following the instructions in Table \ref{rating_instruction}.

\begin{table*}[htbp]
\centering
\begin{tabular}{p{5cm} p{3.5cm} p{3.5cm} p{3.5cm}}
\toprule
\textbf{Criterion and Explanation} & \textbf{1 - 3 (Very Bad)} & \textbf{3 - 5 (OK, but not good enough)} & \textbf{5 - 7 (Good to Very Good)} \\
\midrule
\textbf{Coverage:} The explanation contains important, salient information and does not miss any important points that contribute to the label prediction. & The explanation misses the most critical points in the input text. & The explanation provides a reason for the prediction, but not the main reason. & The explanation covers the most important points/reasons for the prediction. \\
\midrule
\textbf{Non-redundancy:} The explanation does not contain any information that is redundant, repeated, or irrelevant to the claim and predicted label. It should also be reasonable according to common sense. & The explanation contains irrelevant information, unnecessary repetition, or elements that do not appear in the input text; violates common sense. & The explanation is acceptable but contains some redundancy or repetition. & Slightly to no redundancy, repetition, or hallucination. \\
\midrule
\textbf{Non-contradiction:} The explanation does not contain any pieces of information that are contradictory to the predicted label and the input text. & The explanation contradicts the predicted label or input text; they address different topics. & The explanation matches the predicted label but is not fully logical. & The explanation and predicted label are fully consistent and logical. \\
\midrule
\textbf{Overall Quality:} Rank the explanations by their overall quality. Consider grammar, readability, and clarity. & Many grammatical errors, difficult to understand. & No major grammar mistakes, but not easy to understand. & Perfect grammar and language clarity. \\
\bottomrule
\end{tabular}
\caption{Rating Criteria for Generated Natural Language Explanations\label{rating_instruction}}
\end{table*}

\subsection{Pairwise agreement for human annotations}
\label{sec:appendix:annotator agreement}

\begin{table}[h]
\centering
\scalebox{0.55}{
\begin{tabular}{c|cc|cc|cc|ccc}\toprule
- &\multicolumn{2}{c}{\textbf{Coverage}} &\multicolumn{2}{c}{\textbf{Non-redundancy}} &\multicolumn{2}{c}{\textbf{Non-contradiction}} &\multicolumn{2}{c}{\textbf{Overall}} \\\cmidrule{2-9}
\textbf{Annotator\_id} &\textbf{2} &\textbf{3} &\textbf{2} &\textbf{3} &\textbf{2} &\textbf{3} &\textbf{2} &\textbf{3} \\\midrule
\multicolumn{9}{c}{\textbf{e-SNLI}} \\
\arrayrulecolor{black!30}\midrule
\textbf{1} &0.51 &0.25 &0.53 &0.43 &0.36 &0.19 &0.33 &0.16 \\
\textbf{2} &- &0.40 &- &0.53 &- &0.43 &- &0.37 \\
  \arrayrulecolor{black!10}\midrule
\textbf{Mean} &\multicolumn{2}{c}{0.39} &\multicolumn{2}{c}{0.49} &\multicolumn{2}{c}{0.33} &\multicolumn{2}{c}{0.29} \\
  \arrayrulecolor{black!70}\midrule
\multicolumn{9}{c}{\textbf{ECQA}} \\
  \arrayrulecolor{black!30}\midrule
\textbf{1} &0.35 &0.20 &0.33 &0.15 &0.58 &0.40 &0.27 &-0.02 \\
\textbf{2} &- &0.10 &- &0.29 &- &0.35 &- &0.30 \\
\arrayrulecolor{black!10}\midrule
\textbf{Mean} &\multicolumn{2}{c}{0.22} &\multicolumn{2}{c}{0.26} &\multicolumn{2}{c}{0.44} &\multicolumn{2}{c}{0.18} \\
\arrayrulecolor{black!100}\bottomrule
\end{tabular}}
\caption{Pairwise agreement for human annotations on e-SNLI and ECQA. We report separately the agreement between annotator pairs 1-2,  2-3, and 1-3. Mean represents the average over three pairwise agreements. }\label{tab:agreemeent}
\end{table}

Table \ref{tab:agreemeent} shows Pairwise agreement for human annotations for NLE generated by T5-based \method on e-SNLI and ECQA dataset. 

\end{document}